\documentclass[nohyperref]{article}

\usepackage{microtype}
\usepackage{graphicx}
\usepackage{booktabs} % for professional tables
\usepackage{amssymb,bm,amsthm}
\usepackage{enumerate}
\usepackage{color, colortbl}
\usepackage{booktabs, multirow}
\definecolor{darkgreen}{rgb}{0, 0.5, 0}
\definecolor{red}{rgb}{1, 0, 0}
\definecolor{purple}{rgb}{0.5, 0, 0.5}
\usepackage{chngpage}
\usepackage{comment}
\usepackage{mfirstuc}
\usepackage{mathtools}
\usepackage{lettrine}

\newcommand\ie{\textit{i.e.,\xspace}}
\newcommand\eg{\textit{e.g.,}}

\newcommand\etc{\textit{etc.}}

\newcommand{\beq}{\begin{equation}}
\newcommand{\eeq}{\end{equation}}
\newcommand{\beqnn}{\begin{equation*}}
\newcommand{\eeqnn}{\end{equation*}}
\newcommand{\beqy}{\begin{eqnarray}}
\newcommand{\eeqy}{\end{eqnarray}}
\newcommand{\beqynn}{\begin{eqnarray*}}
\newcommand{\eeqynn}{\end{eqnarray*}}
\newcommand{\bit}{\begin{itemize}}
\newcommand{\eit}{\end{itemize}}
\newcommand{\ben}{\begin{enumerate}}
\newcommand{\een}{\end{enumerate}}
\newcommand{\bex}{\begin{example}}
\newcommand{\eex}{\end{example}}

\newcommand{\balg}[1]{\begin{algorithm} \caption{#1}}
\newcommand{\ealg}{\end{algorithm}}

\newcommand{\balgc}{\begin{algorithmic}[1]}
\newcommand{\ealgc}{\end{algorithmic}}

\newcommand{\bary}{\begin{array}}
\newcommand{\eary}{\end{array}}
\newcommand{\bmx}{\begin{bmatrix}}
\newcommand{\emx}{\end{bmatrix}}
\newcommand{\bsmx}{\left[\begin{smallmatrix}}
\newcommand{\esmx}{\end{smallmatrix}\right]}
\newcommand{\bmxc}[1]{\left[\begin{array}{@{}#1@{}}}
\newcommand{\emxc}{\end{array}\right]}
\newcommand{\bcn}{\begin{center}}
\newcommand{\ecn}{\end{center}}

\usepackage{amsmath,amsfonts,bm}

\def\1{\bm{1}}

\DeclareMathAlphabet{\mathsfit}{\encodingdefault}{\sfdefault}{m}{sl}
\SetMathAlphabet{\mathsfit}{bold}{\encodingdefault}{\sfdefault}{bx}{n}

\usepackage{subfig}
\usepackage{floatrow}
\usepackage{wrapfig}
\usepackage[colorlinks=true, linkcolor=black, citecolor=darkblue]{hyperref}
\usepackage{xspace}
\usepackage{multirow, bigstrut}
\usepackage{amsmath}

\newtheorem*{theorem*}{Theorem}

\usepackage[accepted]{icml2022}

\usepackage{amsmath}
\usepackage{amssymb}
\usepackage{mathtools}
\usepackage{amsthm}

\usepackage{todonotes}

\usepackage[capitalize,noabbrev]{cleveref}

\theoremstyle{plain}
\theoremstyle{definition}
\theoremstyle{remark}

\icmltitlerunning{Graph Neural Networks Intersect Probabilistic Graphical Models: A Survey}

\begin{document}

\twocolumn[
\icmltitle{Graph Neural Networks Intersect Probabilistic Graphical Models:\\ A Survey }

% sumathbfissions: the style file will automatically remove it for you


\icmlsetsymbol{equal}{*}

\begin{icmlauthorlist}
\icmlauthor{Chenqing Hua}{yy,yyyy}
\icmlauthor{Sitao Luan}{yy,yyyy}
\icmlauthor{Qian Zhang}{yyy}
\icmlauthor{Jie Fu}{yyyyy}
\end{icmlauthorlist}

\icmlaffiliation{yy}{McGill University}
\icmlaffiliation{yyy}{Shandong Agricultural University}
\icmlaffiliation{yyyy}{Mila}
\icmlaffiliation{yyyyy}{BAAI}

\icmlcorrespondingauthor{Chenqing Hua}{Chenqing.hua@mail.mcgill.ca}

%\icmlkeywords{Machine Learning, ICML}

\vskip 0.3in
]

\printAffiliationsAndNotice{} %


%\printAffiliationsAndNotice{}  % leave blank if no need to mention equal contribution % otherwise use the standard text.

\begin{abstract}
Graphs are a powerful data structure to represent relational data and are widely used to describe complex real-world data structures. Probabilistic Graphical Models (PGMs) have been well-developed in the past years to mathematically model real-world scenarios in compact graphical representations of distributions of variables. 
Graph Neural Networks (GNNs) are new inference methods developed in recent years and are attracting growing attention due to their effectiveness and flexibility in solving inference and learning problems over graph-structured data. These two powerful approaches have different advantages in capturing relations from observations and how they conduct message passing, and they can benefit each other in various tasks. In this survey, we broadly study the intersection of GNNs and PGMs. Specifically, we first discuss how GNNs can benefit from learning structured representations in PGMs, generate explainable predictions by PGMs, and how PGMs can infer object relationships. Then we discuss how GNNs are implemented in PGMs for more efficient inference and structure learning. In the end, we summarize the benchmark datasets used in recent studies and discuss promising future directions.
\end{abstract}

\section{Introduction}
\label{sec:introduction}
\begin{figure}[htbp]
\centering
{
\includegraphics[width=1\textwidth]{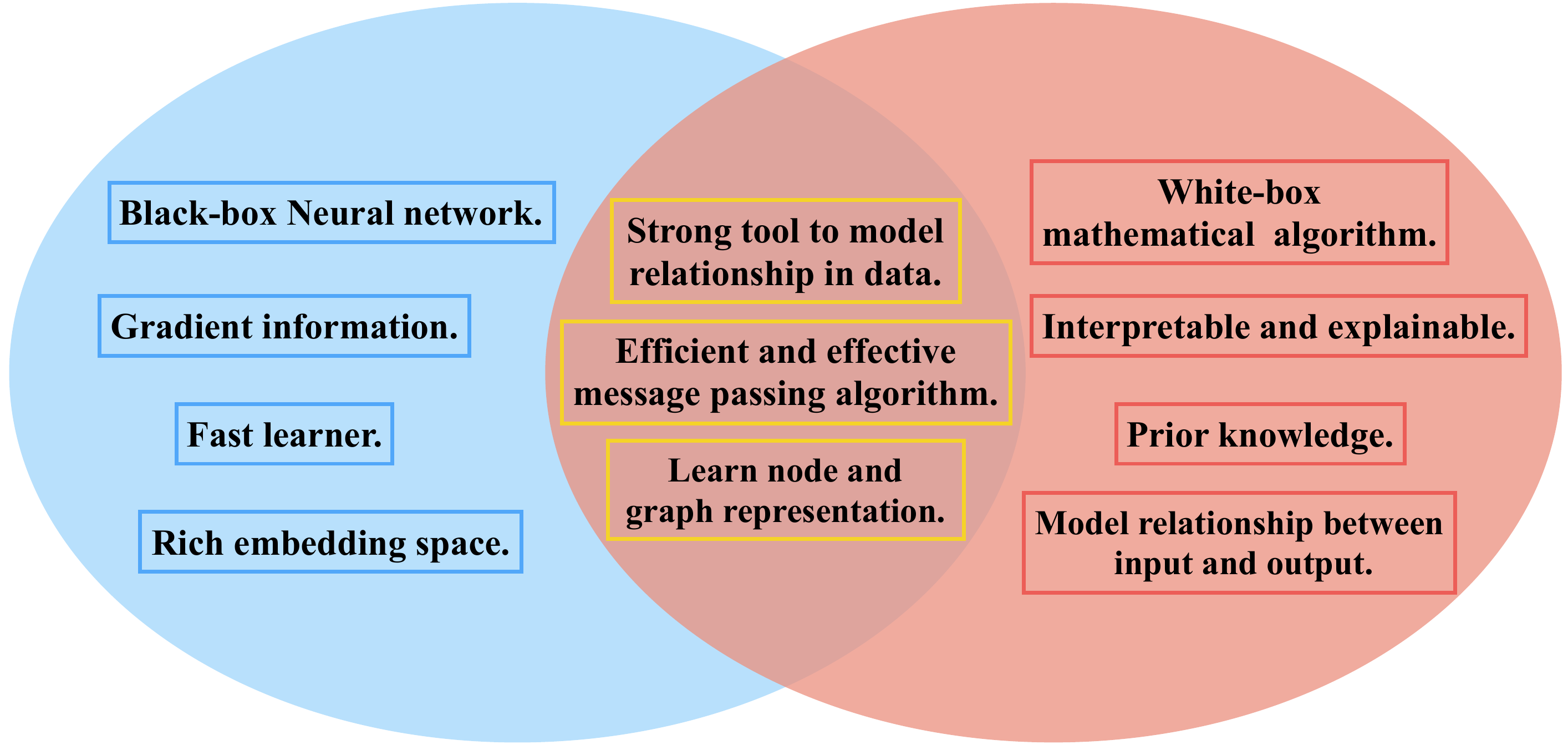}}
{%
  \caption{Visualization of relationships between probabilistic graphical models and graph neural networks. The blue circle concludes the properties of GNNs, the red circle shows the properties of PGMs, and the intersection area shows the shared properties of the two powerful graph learning tools}
  \label{fig:relation}
}
\end{figure}

\begin{figure*}[htbp]
\centering
{
\includegraphics[width=1\textwidth]{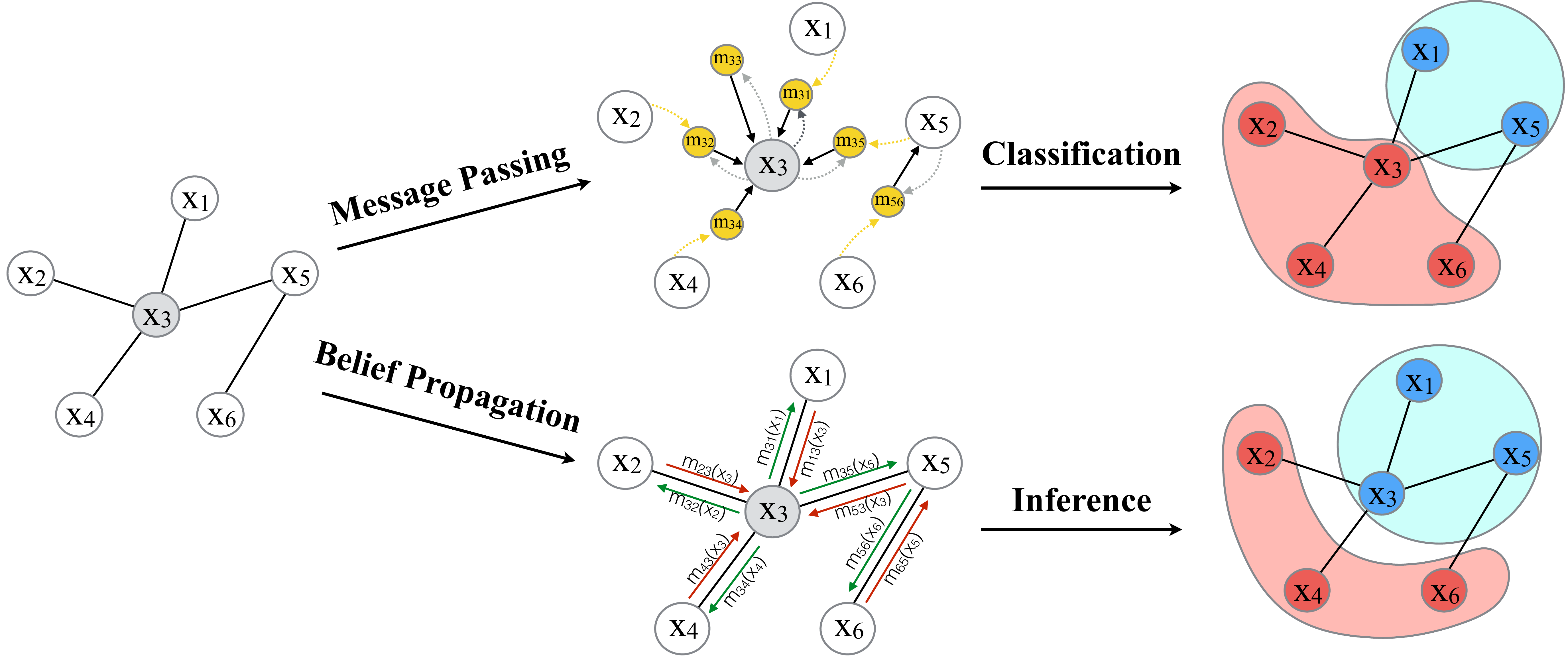}}
{%
  \caption{Illustration of one connection between probabilistic graphical models and graph neural networks on identifying node types, \ie{} exact inference for PGMs or node classification for GNNs. 
  The PGM model executes the belief propagation algorithm on passing messages between nodes and the GNN model executes the message passing algorithm to pass node information to the neighbors. 
  The two powerful algorithms result in different predictions due to their unique message-passing mechanisms.}
  \label{fig:node.task}
}
\end{figure*}

Graphs are expressive to represent objects and their relations with node and edge representations. 
Graph Neural Networks (GNNs)~\citep{kipf2016classification,velivckovic2017attention,hamilton2017inductive, luan2020complete,luan2021heterophily, hua2022high, luan2022revisiting} and Probabilistic Graphical Models (PGMs)~\citep{koller2009probabilistic} are two different powerful tools for learning graph representations and both have been successfully developed to address practical issues when it comes to graph-structured applications. 
GNNs are deep learning architectures that are specifically designed for relational data, which are a generalization of message-passing neural networks~\citep{gilmer2017neural}
for data in non-Euclidean domain. PGMs are mathematically and statistically interpretable models which can express the conditional dependence structure between random variables~\citep{qu2019gmnn,qu2021neural}.
GNNs and PGMs are capable of conducting accurate and fast inferences on graphs but the major differences between them are how they process relational variables and how they conduct message passing in the models, which are shown in Figure~\ref{fig:relation} and \ref{fig:node.task}.

Figure~\ref{fig:relation} summarizes the common and distinct properties of GNNs and PGMs, \eg{} GNNs produce richer representations because they project features onto an informative embedding space while PGMs generate more concrete representations because they always model the relationships between input data and output representations. 
Figure~\ref{fig:node.task} demonstrates how GNNs and PGMs perform differently  when they conduct message passing/belief propagation on a node classification/exact inference task.
%Figure~\ref{fig:node.task} illustrate the message-passing connection when a GNN performs node classifiction/PGM performs exact inference. 
Generally speaking, the two methods have some unique advantages, and the similarities allow the two models to be incorporated together for more efficient and effective algorithms. 
We summarize five main directions where they can benefit each other.

\textbf{PGMs Benefit GNNs}

The first direction of GNNs is to achieve structured representations and predictions via \textit{Conditional Random Fields} (CRF).
The problem arises because most GNNs hold a strong assumption for which the node representations and labels are independent when node features and edges are observed~\citep{qu2021neural}, while PGMs always try to model the relationships between input data and output representations~\citep{qu2019gmnn}. It is argued that the dependence between input data and output predictions should not be ignored in a GNN~\citep{gao2019conditional, qu2019gmnn, qu2021neural}. To achieve structured predictions, GNNs can adopt CRFs in PGMs, which can further address the independence issue. We will discuss CRFs enhanced GNNs that model the dependence between node input representations and embeddings in Section~\ref{sec:sec4.1}. 

More than achieving structured representations and predictions for GNNs through CRFs, the second direction is that other PGM methods can be mixed with GNNs for bette classification results in terms of the algorithm accuracy and efficiency. We will discuss other PGM refinement methods in Section~\ref{sec:4.2.Refinements}. 

The second direction is the explainability of GNNs. Since complex graph structures and node features always lead to complex composite of GNN architectures, GNN predictions cannot be often easily explained~\citep{ying2019gnnexplainer}.
However, we usually desire to give reasonable explanations to those hard predictions~\citep{ying2019gnnexplainer, yuan2020explainability}. 
And white-box PGMs can provide an opportunity to generate reasonable graphical explanations for those predictions. We will discuss PGMs for explainable GNNs in Section~\ref{sec:4.3.explainable}. 

The third direction is graph structure learning for GNNs. The question comes when people demand well-defined graph structures for GNNs to perform various tasks but the structures are not well-defined. Graph structure learning is always important when the input graph structure is noisy or sometimes missing~\citep{kipf2018neural,zhu2021deep}, the relations are not well-defined, the edges do not reflect exact relationships~\citep{wang2020learning}. PGMs can effectively model and infer object relationships from observed variables based on pairwise object similarities or the potential that an edge would appear between two objects so that they can assist GNNs to learn graph structures. We will discuss PGMs enhanced graph structure learning in Section~\ref{sec:4.4.structure}.

\textbf{GNNs Benefit PGMs}

The first direction is that GNNs can help PGMs for better inference through more efficient message-passing algorithms. As illustrated in figure~\ref{fig:node.task}, both PGMs and GNNs conduct message-passing when aggregating information in a neighborhood. However, \textit{Loopy Belief Propagation} (LBP) algorithm~\citep{murphy2013loopy} is relatively slower than message passing used in GNNs and will normally fail on real-world loopy networks~\citep{murphy2013loopy, yoon2019inference}. Message passing algorithm used in GNNs can be adopted to improve LBP in terms of efficiency and accuracy. We will discuss this direction in Section~\ref{sec:5.1.inference}. 

The second direction is that GNNs can benefit \textit{Directed Acyclic Graphs} (DAG) learning for PGMs.
Learning a DAG from observed variables is always expensive and is an extremely challenging NP-hard problem~\citep{chickering1996learning}. Nevertheless, GNNs can be implemented to intermediately enhance DAG learning due to their capability of capturing nonlinear relationships.  We will discuss GNNs enhanced DAG learning in Section~\ref{sec:5.2.dag}. 

In a nutshell, PGMs can be used to improve the prediction performance, explainability and structure learning ability of GNNs;  GNNs can be implemented to enhance PGMs for inference and DAG learning. There is no existing literature reviewing the intersection of GNNs and PGMs. Therefore, in this survey, we formulate the current limitations of GNNs and PGMs that can potentially be addressed by combining each other. This paper is organized as follows: we introduce related background knowledge in Section~\ref{sec:prelimiaries} and \ref{sec:tasks}; we discuss applications of PGMs used in GNNs in Section~\ref{sec:sec3} and discuss applications of GNNs used in PGMs in Section~\ref{sec:sec5}; a comprehensive list of  benchmark datasets is summarized by category in Section~\ref{sec:sec6}; in Section~\ref{sec:sec7}, we outline the challenges and potential research directions for future studies.

\section{Preliminaries}
\label{sec:prelimiaries}

\paragraph{Notations} We use bold font letters for vectors (\eg{} $\mathbf{x}$), capital letters (\eg{} $\mathbf{X}$) for matrices, and regular ones for nodes and edges (\eg{} $v,e$). Let  ${G}=(V,{E})$ be an undirected, unweighted graph where $V=\{v_1,...,v_n\}$ is the node set and $E\in{V\times V}$ is the edge set, $v_i\in V$ to denote the a node and $e_{ij}\in E$ to denote an edge associated with nodes $v_i,v_j$. The neighborhood of a node $v$
is defined as $N(v) = \{u\in V \ | \ e_{vu}\in E\}$. The adjacency
matrix $A$ is a $N\times N$ matrix with $A_{ij}=1$ if $e_{ij}\in E$ and $A_{ij}=0$ otherwise. $\mathbf{X} \in \mathbb{R}^{n\times F}$ denotes the node feature matrix, where $\mathbf{x}_i\in \mathbb{R}^F$ corresponds to the feature of node $v_i$. Each node $v_i$ is associated with a discrete label or a continuous value $\mathbf{y}_{v_i}$.

\paragraph{Graph Neural Networks} \textit{Graph Neural Networks} (GNN) can learn rich node representations on large-scale networks and
achieve state-of-the-art results in node classification and edge prediction, so they have gradually
become an important tool of conducting network analysis and related tasks.
Message-passing GNNs iteratively update node representations in a learning task. Given the input of the first layer $\mathbf{H}^{(0)}=\mathbf{X}$, the learned node hidden representations at each $l$-th layer of a GNN can be denoted as $\mathbf{H}^{(l)}=[\mathbf{h}^{(l)}_v]_{v\in V}$, and the node representations $\mathbf{H}^{(L)}$ from the last GNN layer $L$ can be fed into a classifier network for predicting classes or relevant properties. For graph classification, an additional  $\mbox{READOUT}(\cdot)$ function aggregates node representations from the final layer to obtain a graph representation $\mathbf{h}_G$ of graph $G$ as,
$$
\mathbf{h}_G = \mbox{READOUT}(\mathbf{H}^{(L)}),
$$
where $\mbox{READOUT}(\cdot)$ can be  be a simple permutation invariant function (\eg{} sum, mean, max pooling, \etc{}).

\paragraph{Message Passing in Graph Neural Networks} GNNs encode node features along with local graph structures through a 
\textit{message-passing mechanism}, by iteratively propagating the messages from local neighbor
representations, and then combining the aggregated messages with node themselves~\citep{hamilton2020graph}. Formally, at the $l$-th layer of a GNN, the message-passing mechanism follows:
\begin{equation}
\label{eq:eq.mp.gnn}
    \begin{aligned}
    \mathbf{m}^{(l)}_{v}&= \mbox{AGGREGATE}^{(l)}(\{\mathbf{h}^{(l-1)}_u,\forall u\in N(v)\}), \\ 
    \mathbf{h}_v^{(l)} &= \mbox{UPDATE}^{(l)}(\mathbf{h}^{(l-1)}_v, \mathbf{m}^{(l)}_{v}),
   \end{aligned}
\end{equation}
where $\mbox{AGGREGATE}^{(l)}(\cdot)$ and $\mbox{UPDATE}^{(l)}(\cdot)$ are permutation invariant aggregation function (\eg{} mean, sum, max pooling) and update funtion (\eg{} linear-layer combination). In words, $\mbox{AGGREGATE}^{(l)}(\cdot)$ first aggregates information from ${N}(v)$, then $\mbox{UPDATE}^{(l)}(\cdot)$ combines the aggregated neighborhood message $\mathbf{m}^{(l)}_{v}$ and previous node representation to give a new representation $\mathbf{h}^{(l)}_{v}$. 
\vspace{-0.2cm}
%\paragraph{Knowledge Graphs} A \textit{knowledge graph} (KG) is a collection of relational facts, each can be represented as a triplet $(h,r,t)$, or a tuple $(E, R, O)$ consisting of an entity set $E$, a relation set $R$, a set of observed triplets $(h,r,t)$.  Each triplet $(h,r,t)$ is associated with a binary variable $\mathbf{v}_{(h,r,t)}$, which $\mathbf{v}_{(h,r,t)}=1$ means true and $0$ otherwise. Given some true observed facts $\mathbf{v}_O=\{\mathbf{v}_{(h,r,t)}=1\}_{(h,r,t)\in O}$, we want to predict valued of unobserved triplets $U$, \ie{} $\mathbf{v}_U=H=\{\mathbf{v}_{(h,r,t)}\}_{(h,r,t)\in H}$.
%In the first-order logic language, entities are \textit{constants} (\eg{} a person or an object), relations are \textit{predicates}. Each predicate $r$ is a logic function defined over $H$, \ie{} $r(\cdot):H\times\dots\times H\rightarrow\{0,1\}$.
\paragraph{Bayesian Networks} \textit{Probabilistic Graphical Models} (PGM) simplify a joint probability
distribution $p(\mathbf{x})$ over many variables $\mathbf{x}$ by factorizing the
distribution according to conditional independence relationships. A \textit{Bayesian Network} (BN) is a \textit{probabilistic graphical model} (PGM) that measures the conditional dependence structure of a set of random variables. % based on the Bayes theorem, $P(A|B)=\frac{P(B|A)P(A)}{P(B)}$. 
A structure of a BN takes the form of a \textit{Directed Acyclic Graph} (DAG) where the loopy structure is not allowed.
\vspace{-0.3cm}
\paragraph{Markov Random Fields} \textit{Markov Random Fields} (MRF) or \textit{Undirected Probabilistic Graphical Models} (uPGM) are undirected graphs $G$ equipped with a Markov property, that is, %for any two nodes $v,u\in V$, 
a node $v$ is independent of $u$ conditioned on $v$'s neighbors, $N(v)$, implying that knowing %$N(v)$ is sufficient for predicting $v$'s value
the distribution on values is sufficient and the node $v$ can be ignorant of anywhere else in the network.
A \textit{Conditional Random Field} (CRF) is a MRF globally conditioned
on nodes $V$ and node features $\mathbf{X}$, which is powerful to model the pair-wise node relationships. 
A pair-wise CRF that
formalizes the joint label distribution follows:
\begin{equation}
\begin{aligned}
p(\mathbf{Y}  \ | \ \mathbf{X},E) = \frac{1}{Z_\phi(\mathbf{X}, E)} \ &\mbox{exp} \ \{\sum_{v\in V}\phi_v(\mathbf{y}_v,\mathbf{X},E) \\
&+ \sum_{e_{vu}\in E}\phi_{vu}(\mathbf{y}_v,\mathbf{y}_u,\mathbf{X},E) \},
\end{aligned}
\end{equation}
where $Z_\phi(\mathbf{X},E)$ is the partition function that ensures the factorization $p(\mathbf{Y}\ | \ \mathbf{X}, E)$ sum to $1$, $\phi_v(\mathbf{y}_v,\mathbf{X}, E)$ and $\phi_{vu}(\mathbf{y}_v,\mathbf{y}_u, \mathbf{X}, E)$ are unary and pair-wise potential functions defined on each node $v$ and each edge $e_{vu}$, respectively.  The
unary function gives the prediction for each individual data point. The pairwise energy function aims to capture the
correlation between the individual data point and its context to regularize the unary function.  CRFs are able to model the joint dependence of node labels
through pair-wise potential function $\phi_{vu}(\mathbf{y}_v,\mathbf{y}_u, \mathbf{X}, E)$.
\vspace{-0.2cm}
\paragraph{Message Passing in Markov Random Fields} \textit{(Sum-Product) Loopy Belief Propagation} algorithm (LBP)~\citep{murphy2013loopy} is a dynamic programming approach to pass information between nodes for answering conditional probability queries in probabilistic graphical models. For each edge $e_{vu}\in E$, a message $\mathbf{m}_{{u}\rightarrow{v}}(\mathbf{y}_v)$, representing a message of node $u$ sending to node $v$,  is
to repeat the following procedure until convergence or after sufficient iterations:
\begin{equation}
\label{eq:eq4.mrf.mp}
\begin{aligned}
    \mathbf{m}_{{u}\rightarrow{v}}(\mathbf{y}_{{v}}) \propto &\sum_{\mathbf{y}_{{u}}} \{ \mbox{exp} \ (\phi_{{v}}(\mathbf{y}_{{v}}, E) + \phi_{{v}{u}}(\mathbf{y}_{{v}},\mathbf{y}_{{u}}, E)) \\
    &\prod_{{v}'\in N({u})\slash{v}} \mathbf{m}_{{v'}\rightarrow{u}}(\mathbf{y}_{{u}}) \}.
\end{aligned}
\end{equation}

Moreover, the \textit{(Max-Product) Loopy Belief Propagation} algorithm follows: 
\begin{equation}
\label{eq:eq4.mrf.maxmp}
\begin{aligned}
    \mathbf{m}_{{u}\rightarrow{v}}(\mathbf{y}_{{v}}) \propto \ &\underset{{\mathbf{y}_{{u}}}}{\mbox{max}} \{ \mbox{exp} \ (\phi_{{v}}(\mathbf{y}_{{v}}, E) + \phi_{{v}{u}}(\mathbf{y}_{{v}},\mathbf{y}_{{u}}, E)) \\
    &\prod_{{v}'\in N({u})\slash{v}} \mathbf{m}_{{v'}\rightarrow{u}}(\mathbf{y}_{{u}}) \}.
\end{aligned}
\end{equation}

The message $\mathbf{m}_{{u}\rightarrow{v}}(\mathbf{y}_v)$ can be derived by simply multiplying the following items: messages of node $u$'s neighbors (except node $v$) sending to $u$, the unitary potential of node $v$, and the pair-wise potential of edge $e_{vu}$. 

%\paragraph{Markov Logic Networks} A \textit{Markov Logic Network} (MLN) is a MRF that uses  logic formulae to define potential functions. Common logic rules to encode domain knowledge include: (1) \textit{Composition rule}. A relation $r_k$ is a composition of $r_i$ and $r_j$ if  $\forall x,y,z\in E, \mathbf{v}_{(x,r_i,y)}\land\mathbf{v}_{(y,r_j,z)} \Rightarrow\mathbf{v}_{(x,r_k,z)}$, (2) \textit{Inverse rule}. A relation $r_j$ is an inverse of $r_i$ if $\forall x,y\in E, \mathbf{v}_{(x,r_i,y)}\Rightarrow\mathbf{v}_{(y,r_j,x)}$, (3) \textit{Symmetric rule} A relation $r$ is symmetric if $\forall x,y\in E, \mathbf{v}_{(x,r,y)}\Rightarrow\mathbf{v}_{(y,r,x)}$, and (4) \textit{Subrelation rule}. A relation $r_j$ is s a subrelation of $r_i$ if $\forall x,y\in E, \mathbf{v}_{(x,r_i,y)}\Rightarrow\mathbf{v}_{(x,r_j,y)}$.

\section{Tasks}
\label{sec:tasks}
In this section, we introduce the main learning tasks for GNNs and PGMs that are studied in this survey. For GNN-related tasks, we have four main categories including structured prediction for node and graph classification, GNN explanation, graph structure learning; and for PGM-related tasks, we have two main categories including exact inference and DAG learning.
For each category  and task, we summarize the representative models in Table~\ref{tab:tab1}, and introduce the relevant datasets in Table~\ref{tab:tab2}.
\subsection{Graph Neural Network Tasks}
We discuss how PGMs can be applied in GNNs in Section~\ref{sec:sec3}.

\paragraph{(Semi-supervised) Node Classification}
For the node classification task of a graph, we develop effective algorithms to determine the labelling or label distributions of nodes with associated node features by looking at the labels of their neighbors, in which we always model an interconnected set of nodes. % instead of a set of \textit{independent and identically distributed} nodes.
Moreover we often refer node classification as \textit{semi-supervised node classification} for which only a small proportion of nodes are provided with labels but we can still access the features of validation and test nodes and their neighborhood structures for training.

\paragraph{Graph Classification}

For the graph classification task of a set of graphs, we seek to learn label distributions of different graphs with associated graph features but instead of making predictions over the individual components of a single graph. We
are particularly given a set of different graphs and our goal is to make independent predictions
specific to each graph, \ie{} graph classification follows the \textit{independent and identically distributed} assumption.

\paragraph{GNN Explanation}
For black-box GNNs, incorporating both graph structures and feature information leads to complex architectures and their predictions are often unexplainable. And the explanation task aims to generate explanations for GNN predictions that could potentially be understood. This task usually determines which nodes and nodes features are crucial and beneficial for message aggregation and prediction.

\paragraph{Graph Structure Learning}
Graph structure learning aims to jointly learn an optimized graph structure and corresponding graph representations when a given graph is noisy or incomplete, or we want to infer relations from observed data. The learned graphs are usually less noisy and more complete than the given graphs, thus refined optimal structures can be used to better conduct downstream tasks, \eg{} node classification, object classification.

\subsection{Probabilistic Graphical Model Tasks}
We discuss how GNNs can be applied in PGMs in Section~\ref{sec:sec5}.

\paragraph{Exact Inference}
For exact inference in PGMs, we expect to learn probability distributions over large numbers of random variables that interact with each other (\eg{} node label distributions over a set of observed nodes). Moreover, we do not only expect accurate predicted probability distributions and predictions (\eg{} node classification for which the model learns label distributions over nodes), but also want to be able to extract meaningful relationships between the model inputs and outputs.

\paragraph{Directed Acyclic Graph Learning}

The directed acyclic graph learning task aims to learn a faithful directed acyclic graph from observed samples of a joint distribution. Learning DAG is a challenging combinatorial problem and the refined DAGs can be further used in downstream tasks, \eg{} inference. Structural learning of DAGs from observed data plays a vital part in causal inference with many applications in real-world scenarios.

\begin{table*}[htbp]
  \centering
  \caption{Summary of models of GNNs and PGMs in this survey.}
  \resizebox{\textwidth}{!}{
    \begin{tabular}{ccccccccccccccccccccccc}
    \hline
    \hline
    \multicolumn{2}{c}{\multirow{2}[4]{*}{Category}} & \multicolumn{2}{c}{\multirow{2}[4]{*}{Method}} & \multirow{2}[4]{*}{Reference} & \multirow{2}[4]{*}{Section} & \multicolumn{4}{c}{Sturtured Prediction for Classification} & \multicolumn{2}{c}{\multirow{2}[4]{*}{GNN Explanation}} & \multicolumn{3}{c}{\multirow{2}[4]{*}{Graph Structure Learning}} & \multicolumn{4}{c}{MRF Refinments for Classification} & \multicolumn{2}{c}{\multirow{2}[4]{*}{DAG Learning}} & \multicolumn{2}{c}{Inference} \bigstrut\\
\cline{7-10}\cline{16-19}\cline{22-23}    \multicolumn{2}{c}{} & \multicolumn{2}{c}{} &       &       & \multicolumn{2}{c}{Hidden Represnetation} & \multicolumn{2}{c}{Output} & \multicolumn{2}{c}{} & \multicolumn{3}{c}{}  & \multicolumn{2}{c}{LBP} & \multicolumn{2}{c}{BN} & \multicolumn{2}{c}{} & MP    & Factor \bigstrut\\
    \hline
    \multicolumn{2}{c}{} & \multicolumn{2}{c}{GCN-CRF} & \citep{gao2019conditional} & \ref{para:CRF.Structured.Hidden.Representations} & \multicolumn{2}{c}{$\surd$} & \multicolumn{2}{c}{} & \multicolumn{2}{c}{} & \multicolumn{3}{c}{}  & \multicolumn{2}{c}{} & \multicolumn{2}{c}{} & \multicolumn{2}{c}{} & \multicolumn{2}{c}{} \bigstrut[t]\\
    \multicolumn{2}{c}{} & \multicolumn{2}{c}{GCN-HCRF} & \citep{liu2019graph} & \ref{para:CRF.Structured.Hidden.Representations} & \multicolumn{2}{c}{$\surd$} & \multicolumn{2}{c}{} & \multicolumn{2}{c}{} & \multicolumn{3}{c}{}  & \multicolumn{2}{c}{} & \multicolumn{2}{c}{} & \multicolumn{2}{c}{} & \multicolumn{2}{c}{} \\
    \multicolumn{2}{c}{} & \multicolumn{2}{c}{MGCN} & \citep{tang2021mutual} & \ref{para:CRF.Structured.Hidden.Representations} & \multicolumn{2}{c}{$\surd$} & \multicolumn{2}{c}{} & \multicolumn{2}{c}{} & \multicolumn{3}{c}{}  & \multicolumn{2}{c}{} & \multicolumn{2}{c}{} & \multicolumn{2}{c}{} & \multicolumn{2}{c}{} \\
    \multicolumn{2}{c}{} & \multicolumn{2}{c}{CGNF} & \citep{ma2018cgnf} & \ref{para:CRF.Structured.Outputs} & \multicolumn{2}{c}{} & \multicolumn{2}{c}{$\surd$} & \multicolumn{2}{c}{} & \multicolumn{3}{c}{}  & \multicolumn{2}{c}{} & \multicolumn{2}{c}{} & \multicolumn{2}{c}{} & \multicolumn{2}{c}{} \\
    \multicolumn{2}{c}{} & \multicolumn{2}{c}{GMNN} & \citep{qu2019gmnn} & \ref{para:CRF.Structured.Outputs} & \multicolumn{2}{c}{} & \multicolumn{2}{c}{$\surd$} & \multicolumn{2}{c}{} & \multicolumn{3}{c}{}  & \multicolumn{2}{c}{} & \multicolumn{2}{c}{} & \multicolumn{2}{c}{} & \multicolumn{2}{c}{} \\
    \multicolumn{2}{c}{\textbf{Probabilistic}} & \multicolumn{2}{c}{StructPool} & \citep{yuan2020structpool} & \ref{para:CRF.Structured.Outputs} & \multicolumn{2}{c}{} & \multicolumn{2}{c}{$\surd$} & \multicolumn{2}{c}{} & \multicolumn{3}{c}{}  & \multicolumn{2}{c}{} & \multicolumn{2}{c}{} & \multicolumn{2}{c}{} & \multicolumn{2}{c}{} \\
    \multicolumn{2}{c}{\textbf{Graphical Models}} & \multicolumn{2}{c}{SMN} & \citep{qu2021neural} & \ref{para:CRF.Structured.Outputs} & \multicolumn{2}{c}{} & \multicolumn{2}{c}{$\surd$} & \multicolumn{2}{c}{} & \multicolumn{3}{c}{}  & \multicolumn{2}{c}{} & \multicolumn{2}{c}{} & \multicolumn{2}{c}{} & \multicolumn{2}{c}{} \\
    \multicolumn{2}{c}{\textbf{Enhanced}} & \multicolumn{2}{c}{PGM-Explainer} & \citep{vu2020pgm} & \ref{sec:4.3.explainable} & \multicolumn{2}{c}{} & \multicolumn{2}{c}{} & \multicolumn{2}{c}{$\surd$} & \multicolumn{3}{c}{}  & \multicolumn{2}{c}{} & \multicolumn{2}{c}{} & \multicolumn{2}{c}{} & \multicolumn{2}{c}{} \\
    \multicolumn{2}{c}{\textbf{Graph Neural Networks}} & \multicolumn{2}{c}{NRI} & \citep{kipf2018neural} & \ref{sec:4.4.structure} & \multicolumn{2}{c}{} & \multicolumn{2}{c}{} & \multicolumn{2}{c}{} & \multicolumn{3}{c}{$\surd$} & \multicolumn{2}{c}{} & \multicolumn{2}{c}{} & \multicolumn{2}{c}{} & \multicolumn{2}{c}{} \\
    \multicolumn{2}{c}{} & \multicolumn{2}{c}{LGS} & \citep{franceschi2019learning} & \ref{sec:4.4.structure} & \multicolumn{2}{c}{} & \multicolumn{2}{c}{} & \multicolumn{2}{c}{} & \multicolumn{3}{c}{$\surd$} & \multicolumn{2}{c}{} & \multicolumn{2}{c}{} & \multicolumn{2}{c}{} & \multicolumn{2}{c}{} \\
    \multicolumn{2}{c}{} & \multicolumn{2}{c}{BGCN} & \citep{zhang2019bayesian} & \ref{sec:4.4.structure} & \multicolumn{2}{c}{} & \multicolumn{2}{c}{} & \multicolumn{2}{c}{} & \multicolumn{3}{c}{$\surd$} & \multicolumn{2}{c}{} & \multicolumn{2}{c}{} & \multicolumn{2}{c}{} & \multicolumn{2}{c}{} \\
    \multicolumn{2}{c}{} & \multicolumn{2}{c}{vGCN} & \citep{elinas2020variational} & \ref{sec:4.4.structure} & \multicolumn{2}{c}{} & \multicolumn{2}{c}{} & \multicolumn{2}{c}{} & \multicolumn{3}{c}{$\surd$} & \multicolumn{2}{c}{} & \multicolumn{2}{c}{} & \multicolumn{2}{c}{} & \multicolumn{2}{c}{} \\
    \multicolumn{2}{c}{} & \multicolumn{2}{c}{DenNE} & \citep{wang2020learning} & \ref{sec:4.4.structure} & \multicolumn{2}{c}{} & \multicolumn{2}{c}{} & \multicolumn{2}{c}{} & \multicolumn{3}{c}{$\surd$} & \multicolumn{2}{c}{} & \multicolumn{2}{c}{} & \multicolumn{2}{c}{} & \multicolumn{2}{c}{} \\
    \multicolumn{2}{c}{} & \multicolumn{2}{c}{BPN} & \citep{bruna2017community} & \ref{sec:4.2.Refinements} & \multicolumn{2}{c}{} & \multicolumn{2}{c}{} & \multicolumn{2}{c}{} & \multicolumn{3}{c}{}  & \multicolumn{2}{c}{} & \multicolumn{2}{c}{$\surd$} & \multicolumn{2}{c}{} & \multicolumn{2}{c}{} \\
    \multicolumn{2}{c}{} & \multicolumn{2}{c}{LinearBP} & \citep{gatterbauer2017linearization} & \ref{sec:4.2.Refinements} & \multicolumn{2}{c}{} & \multicolumn{2}{c}{} & \multicolumn{2}{c}{} & \multicolumn{3}{c}{}  & \multicolumn{2}{c}{$\surd$} & \multicolumn{2}{c}{} & \multicolumn{2}{c}{} & \multicolumn{2}{c}{} \\
    \multicolumn{2}{c}{} & \multicolumn{2}{c}{LCM} & \citep{wang2021semi} & \ref{sec:4.2.Refinements} & \multicolumn{2}{c}{} & \multicolumn{2}{c}{} & \multicolumn{2}{c}{} & \multicolumn{3}{c}{}  & \multicolumn{2}{c}{$\surd$} & \multicolumn{2}{c}{} & \multicolumn{2}{c}{} & \multicolumn{2}{c}{} \\
    \multicolumn{2}{c}{} & \multicolumn{2}{c}{DAGNN} & \citep{thost2021directed} & \ref{sec:4.2.Refinements} & \multicolumn{2}{c}{} & \multicolumn{2}{c}{} & \multicolumn{2}{c}{} & \multicolumn{3}{c}{}  & \multicolumn{2}{c}{} & \multicolumn{2}{c}{$\surd$} & \multicolumn{2}{c}{} & \multicolumn{2}{c}{} \bigstrut[b]\\
    \hline
          &       & \multicolumn{2}{c}{node-GNN\&msg-GNN} & \citep{yoon2019inference} & \ref{sec:5.1.inference} & \multicolumn{2}{c}{} & \multicolumn{2}{c}{} & \multicolumn{2}{c}{} & \multicolumn{3}{c}{}  & \multicolumn{2}{c}{} & \multicolumn{2}{c}{} & \multicolumn{2}{c}{} & $\surd$ &  \bigstrut[t]\\
    \multicolumn{2}{c}{\textbf{Graph Neural Networks}} & \multicolumn{2}{c}{ Factor-GNN} & \citep{fei2021generalization} & \ref{sec:5.1.inference} & \multicolumn{2}{c}{} & \multicolumn{2}{c}{} & \multicolumn{2}{c}{} & \multicolumn{3}{c}{}  & \multicolumn{2}{c}{} & \multicolumn{2}{c}{} & \multicolumn{2}{c}{} & $\surd$ &  \\
    \multicolumn{2}{c}{\textbf{Enhanced}} & \multicolumn{2}{c}{FGNN} & \citep{zhang2020factor} & \ref{sec:5.1.inference} & \multicolumn{2}{c}{} & \multicolumn{2}{c}{} & \multicolumn{2}{c}{} & \multicolumn{3}{c}{}  & \multicolumn{2}{c}{} & \multicolumn{2}{c}{} & \multicolumn{2}{c}{} &       & $\surd$ \\
    \multicolumn{2}{c}{\textbf{Probabilistic}} & \multicolumn{2}{c}{FG-GNN} & \citep{satorras2021neural} & \ref{sec:5.1.inference} & \multicolumn{2}{c}{} & \multicolumn{2}{c}{} & \multicolumn{2}{c}{} & \multicolumn{3}{c}{}  & \multicolumn{2}{c}{} & \multicolumn{2}{c}{} & \multicolumn{2}{c}{} &       & $\surd$ \\
    \multicolumn{2}{c}{\textbf{Graphical Models}} & \multicolumn{2}{c}{DAG-GNN} & \citep{yu2019dag} & \ref{sec:5.2.dag} & \multicolumn{2}{c}{} & \multicolumn{2}{c}{} & \multicolumn{2}{c}{} & \multicolumn{3}{c}{}  & \multicolumn{2}{c}{} & \multicolumn{2}{c}{} & \multicolumn{2}{c}{$\surd$} & \multicolumn{2}{c}{} \\
          &       & \multicolumn{2}{c}{DeepGMG} & \citep{li2018learning} & \ref{sec:5.2.dag} & \multicolumn{2}{c}{} & \multicolumn{2}{c}{} & \multicolumn{2}{c}{} & \multicolumn{3}{c}{}  & \multicolumn{2}{c}{} & \multicolumn{2}{c}{} & \multicolumn{2}{c}{$\surd$} & \multicolumn{2}{c}{} \\
          &       & \multicolumn{2}{c}{D-VAE} & \citep{zhang2019d} & \ref{sec:5.2.dag} & \multicolumn{2}{c}{} & \multicolumn{2}{c}{} & \multicolumn{2}{c}{} & \multicolumn{3}{c}{}  & \multicolumn{2}{c}{} & \multicolumn{2}{c}{} & \multicolumn{2}{c}{$\surd$} & \multicolumn{2}{c}{}
          \bigstrut[b]\\
    \hline
    \hline
    \end{tabular}%
    }
  \label{tab:tab1}%
\end{table*}%

\section{Probabilistic Graphical Models Enhanced Graph Neural Networks} 
\label{sec:sec3}
\begin{figure}[htbp]
\centering
{
\includegraphics[width=1\textwidth]{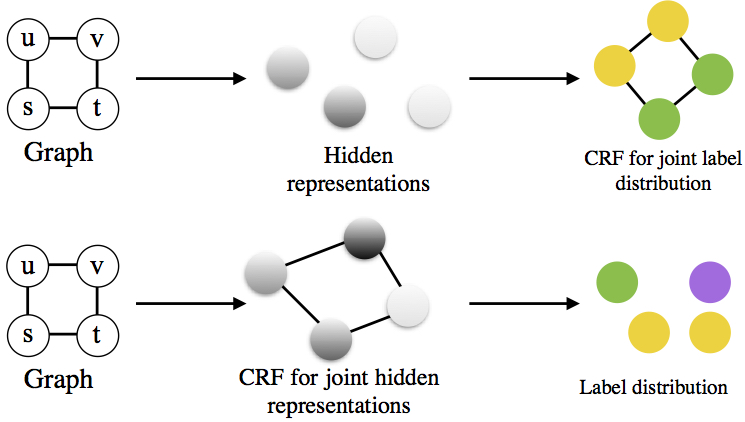}}
{%
  \caption{Overview of CRF implementation for modeling joint dependence of hidden representations and labels in GNN node classification. Top: CRF for structured output (label) prediction. Bottom: CRF for structured hidden representation.}
  \label{fig:crf}
}
\end{figure}

In this section, we present our taxonomy of graph neural networks (GNN) with refinements of probabilistic graphical models (PGM) and Markov random fields (MRF). 
MRFs are generally implemented in GNNs for four purposes: 
(1) CRFs leverage GNNs to achieve \textit{structured representations} for better prediction results; (2) PGM (other than CRF) refinements for GNN classification tasks; (3) PGMs can reason GNNs' predictions for achieving explainable black-box models; and (4) PGMs can be used to infer structures from observed data for GNNs to perform downstream tasks. GNNs with structured representations, in detail, can be categorized into two groups, {structured hidden representations} and {structured outputs} or {structured output predictions}, these two approaches have different initiatives and motivations. 

\subsection{CRFs Enhanced GNNs for Classification}
\label{sec:sec4.1}
Most existing GNNs hold a strong assumption that node hidden representations and node labels are conditionally independent on node features and edges, thus majority of GNN architectures ignore the
joint dependence of node hidden representations and node labels, and the joint node hidden unit distribution is factorized into the marginals as
\begin{equation}
\label{eq:prob.gnn.h}
p(\mathbf{H}^{(l)}  \ | \ \mathbf{X},E) = \prod_{v\in V}p(\mathbf{h}^{(l)}_v \ |\ \mathbf{X},E),
\end{equation}
where each marginal distribution $p(\mathbf{h}^{(l)}_v \ | \ \mathbf{X},E)$ is modeled as a distribution over hidden units.
The joint node label distribution is factorized into the marginals as
\begin{equation}
\label{eq:prob.gnn.y}
p(\mathbf{Y}  \ | \ \mathbf{X},E) = \prod_{v\in V}p(\mathbf{y}_v \ |\ \mathbf{X},E),
\end{equation}
where each marginal distribution $p(\mathbf{y}_v \ | \ \mathbf{X},E)$ is modeled as a categorical distribution over labels.  

However, the assumption might be overstated because we also assume that, in graphs and networks, two connected nodes tend to be similar in terms of their representations and labels~\citep{qu2021neural}, so the joint dependence of node hidden representations and node labels should not be neglected. In equation~\ref{eq:prob.gnn.h}, it cannot guarantee that the obtained hidden representations preserve the similarity relationship explicitly.
In equation~\ref{eq:prob.gnn.y}, the labels of different nodes are separately predicted according to their own marginal label distributions, yet the joint dependence of node labels is ignored.

To overcome the problem, \textit{conditional random fields} (CRF) are implemented in GNN architectures to model the joint dependence of node hidden representations and node labels, and therefore achieve structured representations and predictions.

\paragraph{CRFs for Structured Hidden Representations}
\label{para:CRF.Structured.Hidden.Representations}
CRFs allow GNNs to learn from the joint dependence of node hidden representations to preserve spatial structure information in learning so that the hidden features can explicitly preserve similarity relationships in a local neighborhood.

At every updating iteration, the graph model is transformed into a CRF, where each node $v\in V$ is assigned with a unary potential $\phi_v(\cdot)$ and each edge $e_{vu}\in E$ is assigned with a pair-wise potential $\phi_{vu}(\cdot)$. The unary potential learns an individual node presentation from its local neighborhood information, and the pair-wise potential learns an edge representation from its connecting nodes. And further the pair-wise potentials or edge values can be sent back to each node, \eg{} LBP algorithm. 

At every $l$-th updating iteration, \citep{gao2019conditional} define the unary potential $\phi_v(\mathbf{h}^{(l)}_v, \hat{\mathbf{h}}_{v}^{(l)})$ to enforce node embeddings $\mathbf{H}^{(l)}$
to be close to embeddings, $\hat{\mathbf{H}}_{GNN}^{(l)}$, obtained from a GNN layer. And the pair-wise potential $\phi_{vu}(\mathbf{h}^{(l)}_v,\mathbf{h}^{(l)}_u, \hat{\mathbf{h}}^{(l)}_v,\hat{\mathbf{h}}^{(l)}_u)$ is defined to capture the similarity between two connected nodes in a way two similar nodes tend to have similar hidden representations while two different nodes tend to have different hidden representations.

Authors in \textit{Graph Convolution Network-Hidden Conditional Random Field} (GCN-HCRF)~\citep{liu2019graph} define two unary potentials to enhance learning. The first unary potential $\phi_v(\mathbf{h}_v,\mathbf{x}_v; \theta)$ measures the likelihood of the local feature $\mathbf{x}_v$ assigned to the hidden representation $\mathbf{h}_v$ parameterized by $\theta$, and the second unary potential $\psi_v(\mathbf{m}_v,\mathbf{Y}; \omega)$ measures the compatibility between class labels $\mathbf{Y}$ and hidden representation $\mathbf{m}_v$ parameterized by $\omega$. And the pair-wise potential $\phi_{vu}(\mathbf{h}^{(l)}_v,\mathbf{h}^{(l)}_u, \mathbf{Y})$ measures the compatibility between class labels $\mathbf{Y}$ and a pair of hidden representations $(\mathbf{h}^{(l)}_v,\mathbf{h}^{(l)}_u)$. 

In \textit{Mutual Conditional Random Field-Graph Neural Network} (MCGN)~\citep{tang2021mutual}, a few-shot learning approach, the unary potential $\phi(\mathbf{h}^{(l)}_v, \mathbf{y}_v)$ describes the relationship between the hidden representation $\mathbf{h}^{(l)}_v$ of support
samples and its ground truth label $\mathbf{y}_v$, and the pair-wise potential $\phi(\mathbf{h}^{(l)}_v, \mathbf{h}^{(l)}_u, \mathbf{y}_v, \mathbf{y}_u)$ results in high compatibility when two similar nodes take the same label or when two dissimilar nodes take different labels.

For updating node representations or unary functions, edge values obtained by pair-wise potentials can be propagated back to nodes that connect them until convergence or after sufficient iterations. MRFs are efficient and effective to produce structured hidden representations.

\paragraph{CRFs for Structured Outputs}
\label{para:CRF.Structured.Outputs}
Compared to CRFs applications in modelling structured hidden representations for local representation dependence,
CRFs can directly model the joint node label dependencies and therefore produce structured outputs. After the GNN training, the node representations of a graph can be transformed into a CRF, for which the unary and pair-wise potential functions directly act on label distributions.

\textit{Conditional Graph Neural Fields} (CGNF)~\citep{ma2018cgnf} explicitly model a joint probability of the entire set of node
labels. They define the unary potential $\phi_v(\mathbf{h}_v,\mathbf{y}_v)$ to be the prediction loss of a GNN which measures the compatibility between a node representation $\mathbf{h}_v$ and its label $\mathbf{y}_v$. And the pair-wise potential $\phi_{uv}(\mathbf{y}_v,\mathbf{y}_u, \hat{e}_{vu}, \mathbf{u}_{vu})$ is defined to capture label correlation of two connected nodes which is impacted by the normalized edge weight $\hat{e}_{vu}$ and label correlation weight $\mathbf{u}_{vu}$.

\textit{Graph Markov Neural Networks} (GMNN)~\citep{qu2019gmnn} model the joint label distribution with a CRF, which can
be effectively trained with the variational EM algorithm parameterized by two GNNs. E-step trains a GNN to learn the posterior distributions of node labels, and M-step applies another GNN to learn the pair-wise potential to model the local label dependence. 

In \textit{Structured Graph Pooling} (StructPool)~\citep{yuan2020structpool}, authors consider the graph pooling as a node clustering problem via CRF. They define unary potential $\phi_v(\mathbf{y}_i)$ to measure energy for node $v$ to be assigned to cluster $\mathbf{y}_i$, and pair-wise potential $\phi_{vu}(\mathbf{y}_i,\mathbf{y}_j)$ to be a Gaussian kernel which measures the energy for nodes $v,u$ to be assigned to clusters $\mathbf{y}_i,\mathbf{y}_j$ respectively.

Authors in \textit{Structured Markov Network} (SMN)~\citep{qu2021neural} implement CRFs to achieve structured output predictions for GNN node classification. They parameterize unary potential $\phi_v(\mathbf{y}_v)$ and pair-wise potential $\phi_{vu}(\mathbf{y}_v,\mathbf{y}_u)$ by two GNNs (NodeGNN and EdgeGNN) to separately calculate label distributions over nodes and edges, and run loopy belief propagation on variables for sufficient iterations to obtain joint node label distribution over the node set.  

Since CRFs can successfully model the relations between variables through their pair-wise potential functions, they are a good method of leveraging GNN architectures to calculate the joint node label distribution and further achieve the structured prediction.

\subsection{GNNs with PGM Refinement Methods}
\label{sec:4.2.Refinements}
In the previous Section~\ref{sec:sec4.1}, we introduce CRFs enhanced GNNs for learning through structured representations and predictions, more than that, alternative PGM methods can incorporate with GNNs for better predictions. 

Authors in~\citep{bruna2017community} incorporate \textit{Graph Neural Networks with Belief Propagation} (BPN) to solve semi-supervised node classification. Their architecture consists of a classification network and a propagation procedure. The first step is to use an independent classifier to compute the prior label distributions of nodes without any neighboring information which can be directly used in classification. And the second diffusion step propagates the priors by the LBP algorithm and computes approximate beliefs of nodes to solve node classification.

A \textit{pair-wise Markov Random Field} (pMRF) associates a discrete random variable with each node to model its label. pMRFs define
a joint probability distribution of all random variables as the
product of the node potentials and edge potentials. The edge potential for an edge $e_{vu}$ is called \textit{coupling matrix}, whose $(i,j)$-th entry indicates the likelihood that nodes $u,v$ have labels $\mathbf{y}_i,\mathbf{y}_j$ respectively.

Authors in~\citep{gatterbauer2017linearization} propose a method that computes a closed-form solution of \textit{loopy belief propagation} of any \textit{pairwise Markov random fields} (LinearBP). They assume a \textit{heuristics-based constant coupling matrix} for all
edges and present a universal implicit linear equation system to represent any pMRFs with node potentials. The solution of LBP algorithm efficiently and iteratively converges to an implicit definition of the final beliefs (or marginal node label distributions). LinearBP is further applied to graph models to efficiently solve semi-supervised node classification.

Instead of using a \textit{heuristics-based constant coupling matrix} for all
edge, authors in~\citep{wang2021semi} leverages LinearBP with the ability to learn edge potentials and propose an effective method to calculate marginal node label distributions over any \textit{pair-wise Markov Random Fields} (pMRF). They define an optimization framework to iteratively update and learn the \textit{coupling matrix} (LCM) for every edge. The final \textit{coupling matrix} is applied to LinearBP to obtain the marginal node label distributions for semi-supervised node classification.

More than defining graph models through \textit{pair-wise Markov random fields}, authors in~\citep{thost2021directed} incorporate GNNs with \textit{Bayesian networks} and propose a \textit{directed acyclic graph neural network} (DAGNN). The messaging-passing mechanism of DAGNN follows the flow defined by the partial order by the DAG. In particular, they generalize a graph model into a DAG, and the DAG order allows a GNN to sequentially update node representations based on those of all their predecessors. The messages landed on a node are no longer limited by the multi-hop local neighborhood, and long-range dependencies are preserved in the message passing for better node representations.

%Authors in~\citep{bruna2017community} propose spectral approximations of BP with GNNs to solve the community detection problem.

\subsection{Explainable GNNs}
\label{sec:4.3.explainable}
In this section, we introduce PGMs that benefit explainable GNNs. \citep{yuan2020explainability} gives a comprehensive survey for explainability in GNNs.
Incorporating both graph structures and node features always leads to complex GNN designs and their predictions are almost impossible to be understood and explained. People aim to generate explanations, which could potentially reason message-passing and node representations, for GNN predictions and outputs. Probabilistic graphical models and Markov random fields for graph learning are always interpretable and explainable thus can be implemented to approximately analyze and give reasons to GNN predictions.

\textit{Probabilistic Graphical Model-Explainer} (PGM-Explainer)~\citep{vu2020pgm} identifies crucial graph components for a GNN prediction and generates an explanation in form of a \textit{Bayesian network} (BN). PGM-Explainer outputs conditional probabilities according to varied contributions of graph components.
Given an input graph $G$ and a GNN prediction to be explained, PGM-Explainer generates a perturbed graph $G'$ by randomly perturbing node features of some random nodes. Then
for any node in $G$, PGM-Explainer
records a random variable indicating whether its feature is perturbed in $G$ and its impact on the GNN predictions of a perturbed graph $G'$. To create a local dataset $D$ for sufficient random variables, the procedure is repeated multiple times for various perturbed graphs. Then PGM-Explainer obtains Markov-blankets\footnote{The Markov-blanket defines the boundaries of a system for variables in a network. This means that a subset that contains all the useful information is called a Markov-blanket, and a Markov-blanket is sufficient to infer a random variable with a set of variables. If a Markov-blanket is minimal, it cannot drop any variable without losing information.} to reduce the size of the local dataset $D$. Finally, an interpretable BN is learned to fit the local dataset and explain the predictions of the original GNN model. PGM-Explainer can be used to explain both node
classification and graph classification tasks.

%\subsection{MLNs for Graph Logic Reasoning}

\subsection{Graph Structure Learning with PGMs}
\label{sec:4.4.structure}

\begin{figure}[htbp]
\centering
{
\includegraphics[width=1\textwidth]{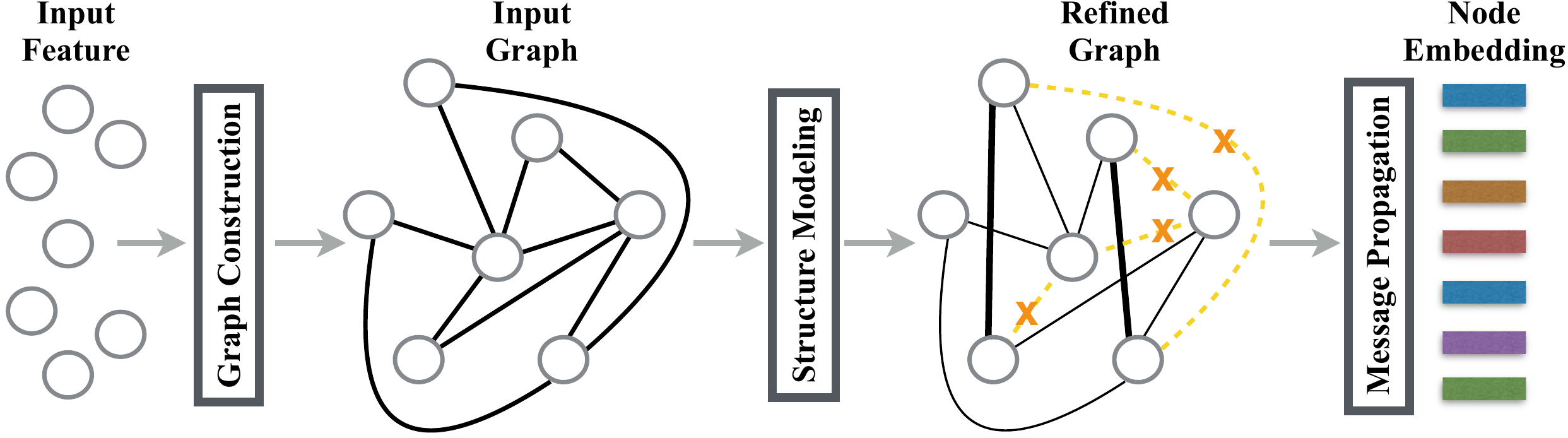}}
{%
  \caption{General illustration of Graph Structure Learning (GSL). GSL models begin with input features and the (optional) graph structure, then the structure modeling module iteratively updates and refines the graph structure (by removing old edges and adding new edges). Finally, node and graph embeddings are obtained from the new refined graph.}
  \label{fig:gsl}
}
\end{figure}

Real-world networks might partially contain noisy structures and edges, or a graph that describes observed data is missing in some applicants. A complete and noiseless graph needs to be learned and inferred from observed samples; figure~\ref{fig:gsl} shows the general pipeline of learning a refined graph from observed data.
PGMs are beneficial to infer and learn graph structures from observed variables for GNNs to conduct downstream tasks (\eg{} node classification, graph classification, \etc{}) when the input graph is unavailable or the given graph is noisy. The core of this task is to model the connectivity of two observed samples. The survey~\citep{zhu2021deep} concludes probabilistic graphical modeling approaches for structure learning to benefit GNNs.

\citet{kipf2018neural} introduce an unsupervised model,  \textit{neural relational inference} (NRI), that learns to infer a graph from observed variables while simultaneously learning the dynamics purely from the observations. NRI takes the form of a variational auto-encoder, in which the encoder maps the observed variables to latent representations of an underlying interaction graph (adjacent matrix), and given the interaction graph, the decoder learns the parameters of a dynamical GNN.

\citet{franceschi2019learning} propose a bi-level programming problem which \textit{learns a discrete and sparse dependence structure} (LGS) of observed variables. LGS learns a structure by modeling the edges between each pair of variables sampled from a Bernoulli distribution while simultaneously training a GNN. They develop a practical algorithm using hyper-gradient estimation to approximately solve the bi-level problem.

\textit{Bayesian Graph Convolutional Network} (BGCN) \citep{zhang2019bayesian} adopt the Bayesian approach, treating the observed graph as a sample from a parametric family of random graphs. They target the inference of the joint posterior of the random graph parameters and node labels. In particular, they utilize Monte Carlo dropout to sample the GCN parameters for several times on each generated graph.

In~\citep{elinas2020variational}, authors aim to carry out a joint inference over GCN parameters and the graph structure (vGCN). They formulate a joint probabilistic model which considers a prior over the adjacency matrix along with a GCN-based likelihood. And they further develop a stochastic variational inference algorithm to jointly estimate the GCN parameters and the graph posterior. In the absence of graph information, their algorithm can effectively solve semi-supervised node classification.

Authors in \citep{wang2020learning} propose an explicit model that learns noise-free node representations while simultaneously eliminates noises over the graphs (DenNE). They propose a generative model with a graph generator and a noise generator, which incorporates different graph priors and generates graph noise to reconstruct the original graph and noise separately. Then, the two generators are jointly optimized via maximum likelihood estimation with real graphs as observations.

%Authors in~\citep{sun2021graph} propose a novel \textit{variational information bottleneck guided graph structure learning framework} (VIB-GSL), which deduces a variational approximation for irregular graph data to form a tractable \textit{information bottleneck} objective function.

\section{Graph Neural Networks Enhanced Probabilistic Graphical Models} 
\label{sec:sec5}
In Section~\ref{sec:sec3}, we discuss how PGMs can be implemented to refine GNNs for various purposes. %(\ie{} structured representations, PGM refinements, explainable GNNs, graph structure learning). 
Conversely, GNNs can be used in PGMs for two purposes: (1) GNNs can efficiently and effectively leverage inference in PGMs; (2) GNNs can enhance the directed acyclic graph learning in PGMs.

\subsection{GNN Enhanced PGMs for Exact Inference}
\label{sec:5.1.inference}
A fundamental computation for exact inference is to compute
the marginal probabilities of task-relevant variables.
The exact inference task aims to learn the joint probability distributions over observed data through a mathematical algorithm. More than this, we also look for a meaningful and interpretable relationship between model inputs and outputs. Since PGMs and GNNs both work on graph-structured data and message-passing mechanism is essential to them, one can compose  PGMs with GNNs for faster and more effective exact inference.

\textit{Loopy belief propagation} (LBP) does not guarantee convergence and can struggle when the conditional dependence graphs contain loops, or it cannot be easily specified for complex continuous probability distributions. Authors in~\citep{yoon2019inference} propose to use GNNs to learn the message-passing algorithm that solves inference problems with loops. They investigate two mappings between graphical models
and GNNs. The first mapping maps a message $\mathbf{m}_{ij}$ between variables $i,j$ in the graphical model to a node $v$ in the GNN. Nodes $v,u$ are connected in the GNN if their corresponding messages share one variable, \eg{} $\mathbf{m}_{ij}$ and $\mathbf{m}_{ik}$. The second mapping maps variables in the probabilistic graphical model to nodes in the GNN and does not provide any hidden states to update the factor nodes. Message-passing algorithms conducted on both GNNs are superior to LBP in PGMs for inference.

\citet{fei2021generalization} propose to learn an iterative message-passing algorithm using GNNs to achieve fast approximate inference on higher-order graphical models that involve many-variable interactions. Their message-passing algorithm performed on GNNs overcomes the drawbacks of BP when a high-order real-world graphical model contains numerous loops.

Moreover GNNs can be generalized into factor graphical models so that not only pairwise dependencies are preserved in the model, but also high-order dependencies of distant variables are kept in the model. 
The factor graphical models can be trained with LBP algorithm to achieve efficient and effective message passing among variables.

\textit{Factor Graph Neural Network}
(FGNN)~\citep{zhang2020factor} generalize a GNN into a factor graphical model which is effective to capture long-range dependencies among multiple variables. In particular, a factor node $\mathbf{f}_c$ is associated with a set of random variables $\mathbf{x}_c$ of a neighborhood of nodes in a graph model. A factor graph model $G=(V, F, E)$ is an analogy to a graph defined in Section~\ref{sec:prelimiaries} but with a set of factor nodes $F$, an additional group of factor features, and extra edges between factors in $F$ and their associated nodes in $V$. FGNN can exactly parameterize the \textit{(Max-Product) loopy belief propagation algorithm} for an effective inference.

\citet{satorras2021neural} also extend graph neural networks to factor graphical models (FG-GNN), and further propose a hybrid approach, \textit{neural enhanced belief propagation} (NEBP).  Their algotithm conjointly runs a FG-GNN with belief propagation. As a result, FG-GNN leverages the belief propagation with the advantages of GNNs. NEBP is an iterative updating algorithm. At every NEBP iteration, an FG-GNN takes messages produced by BP and then runs for two iterations and updates the BP messages. The iterative procedure is repeated sufficient times, and the final refined beliefs are used to compute marginals for inference.

\begin{table*}[htbp]
  \centering
  \caption{Summary of selected datasets for GNNs conducting the node- and graph-level tasks.}
  \resizebox{\textwidth}{!}{
    \begin{tabular}{ccccccccccc}
    \hline
    \hline
    Category & Dataset & Reference & Task  & \#Graphs & \#Node & \#Node Feature & \#Edges & \#Edge Feature & \#Class & Metric \bigstrut\\
    \hline
          & \textit{Cora}  & \citep{sen2008collective} & Node  & 1     & 2,708 & 1433  & 5,429 & \_    & 7     & Accuracy \bigstrut[t]\\
          & \textit{Citeseer} & \citep{sen2008collective} & Node  & 1     & 3,327 & 3703  & 4,732 & \_    & 6     & Accuracy \\
    \textbf{Citation} & \textit{Pubmed} & \citep{sen2008collective} & Node  & 1     & 19,717 & 500   & 44,338 & \_    & 3     & Accuracy \\
          & \textit{ogbn-arxiv} & \citep{hu2020ogb} & Node  & 1     & 2,449,029 & 128   & 1,166,243 & \_    & 40    & Accuracy \\
          & \textit{ogbn-mag} & \citep{hu2020ogb} & Node  & 1     & 1,939,743 & 128   & 21,111,007 & \_    & 349   & Accuracy \\
          & \textit{ogbn-papers100M} & \citep{hu2020ogb} & Node  & 1     & 111,059,956 & 128   & 1,615,685,872 & \_    & 172   & Accuracy \bigstrut[b]\\
    \hline
    \textbf{Social} & \textit{Reddit} & \citep{hamilton2017inductive} & Node  & 1     & 232,965 & 602   & 11,606,919 & \_    & 41    & Accuracy \bigstrut[t]\\
          & \textit{REDDIT-BINARY} & \citep{yanardag2015deep} & Graph & 2,000 & 429.61 (avg.) & \_    & 497.75 (avg.) & \_    & 2     & Accuracy \bigstrut[b]\\
    \hline
    \textbf{Language} & \textit{NELL}  & \citep{carlson2010toward} & Node  & 1     & 65,755 & 61278 & 266,144 & \_    & 210   & Accuracy \bigstrut\\
    \hline
    \textbf{Mathematical} & \textit{PATTERN} & \citep{dwivedi2020benchmarkgnns} & Node  & 14,000 & 117.47 (avg.) & 3     & 4749.15 (avg.) & \_    & 2     & Accuracy \bigstrut[t]\\
    \textbf{Modeling} & \textit{CLUSTER} & \citep{dwivedi2020benchmarkgnns} & Node  & 12,000 & 117.20 (avg.) & 7     & 4301.72 (avg.) & \_    & 6     & Accuracy \bigstrut[b]\\
    \hline
          & \textit{ogbn-proteins} & \citep{hu2020ogb} & Node  & 1     & 132,534 & \_    & 39,561,252 & 8     & 112   & ROC-AUC \bigstrut[t]\\
    \textbf{Protein} & \textit{PROTEINS} & \citep{yanardag2015deep} & Graph & 1,113 & 39.06 (avg.) & 4     & 72.81 (avg.) & \_    & 2     & Accuracy \\
          & \textit{DD}    & \citep{yanardag2015deep} & Graph & 1,178 & 284.31 (avg.) & 82    & 715.65 (avg.) & \_    & 2     & Accuracy \\
          & \textit{ogbg-ppa} & \citep{hu2020ogb} & Graph & 158,100 & 243.4 (avg.) & \_    & 2,266.1 (avg.) & 7     & 37    & Accuracy \bigstrut[b]\\
    \hline
    \textbf{Commercial} & \textit{ogbn-products} & \citep{hu2020ogb} & Node  & 1     & 2,449,029 & 100   & 61,859,140 & \_    & 47    & Accuracy \bigstrut\\
    \hline
    \textbf{Chemistry} & \textit{ZINC}  & \citep{dwivedi2020benchmarkgnns} & Graph & 12,000 & 23.16 (avg.) & 28    & 49.83 (avg.) & 4     & \_    & MAE \bigstrut\\
    \hline
          & \textit{ogbg-molhiv} & \citep{hu2020ogb} & Graph & 41.127 & 25.5 (avg.) & 9     & 27.5 (avg.) & 3     & \_    & ROC-AUC \bigstrut[t]\\
    \textbf{Molecule} & \textit{ogbg-molpcba} & \citep{hu2020ogb} & Graph & 437,929 & 26.0 (avg.) & 9     & 28.1 (avg.) & 3     & \_    & AP \\
          & \textit{NCI1}  & \citep{yanardag2015deep} & Graph & 4,110 & 29.87 (avg.) & 37    & 32.3 (avg.) & \_    & 2     & Accuracy \\
          & \textit{MUTAG} & \citep{yanardag2015deep} & Graph & 188   & 17.93 (avg.) & 7     & 19.79 (avg.) & \_    & 2     & Accuracy \bigstrut[b]\\
    \hline
    \textbf{Computer} & \textit{MNIST} & \citep{dwivedi2020benchmarkgnns} & Graph & 70,000 & 70.57 (avg.) & 3     & 564.53 (avg.) & 1     & 10    & Accuracy \bigstrut[t]\\
    \textbf{Vision} & \textit{CIFAR10} & \citep{dwivedi2020benchmarkgnns} & Graph & 60,000 & 117.63 (avg.) & 5     & 941.07 (avg.) & 1     & 10    & Accuracy \bigstrut[b]\\
    \hline
    \hline
    \end{tabular}%
    }
  \label{tab:tab2}%
\end{table*}%

\subsection{GNN Enhanced Directed Acyclic Graph Learning}
\label{sec:5.2.dag}
Learning \textit{directed acyclic graphs} (DAG) from observed variables is an NP-hard problem, owing mainly to the combinatorial acyclicity constraint that is difficult to enforce efficiently~\citep{chickering1996learning}.
GNNs cannot be directly used in structure learning of probabilistic graphical models for inference, but can intermediately enhance the structure learning due to their capability of capturing complex nonlinear mappings.

\begin{figure}[htbp]
\centering
{
\includegraphics[width=1\textwidth]{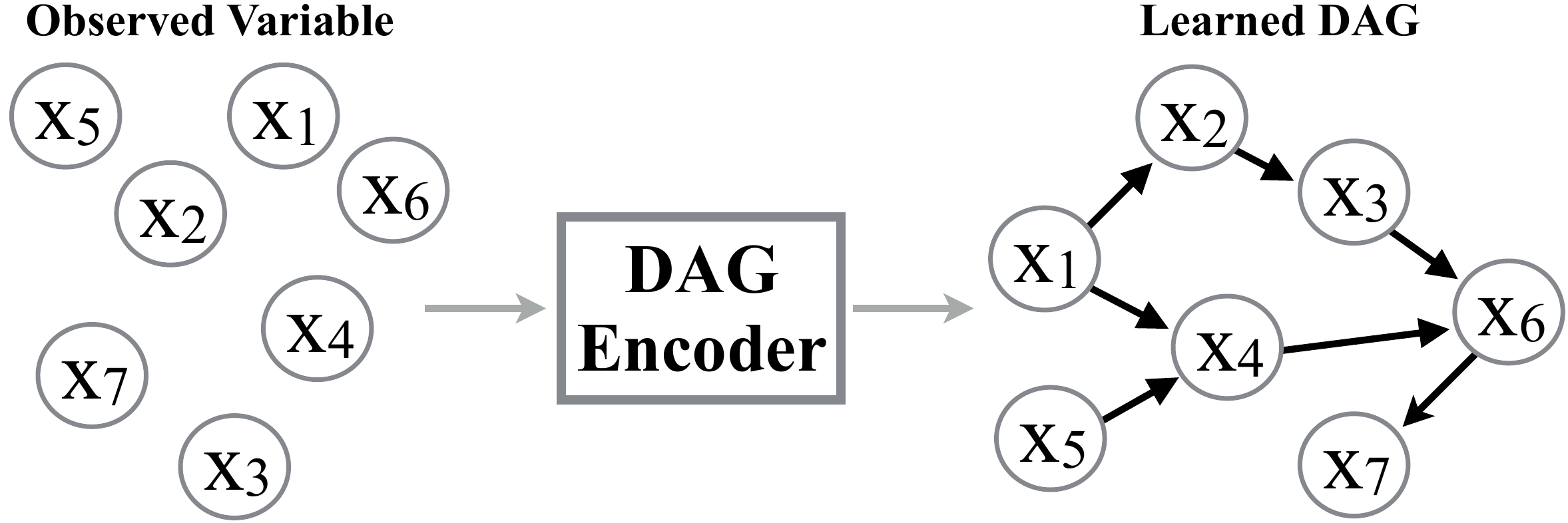}}
{%
  \caption{Overview of DAG learning from observed variables, a DAG structure is generally computed via a DAG autoencoder.}
  \label{fig:crf}
}
\end{figure}

\citet{li2018learning} propose a powerful generative model over graphs (DeepGMG), which can capture both their structures
and attributes. DeepGMG uses a sequential process to generate one node at a time and connects each node to the existing partial graph by creating edges one by one. They use GNNs to express probabilistic dependencies among a graph’s nodes and edges, and can, in principle, learn distributions over any arbitrary graph. Their method can efficiently and effectively embed DAGs by using simultaneous message passing.

\citet{yu2019dag} propose a deep graph-based generative model and apply a variant of the structural constraint to learn a weighted adjacency matrix of a \textit{directed acyclic graph}, in which the generative model is a variational auto-encoder parameterized by a novel \textit{graph neural network} (DAG-GNN). In addition to the rich capacity, an advantage
of the proposed model is that it naturally handles discrete variables as well as vector-valued ones.

\citet{zhang2019d} introduce a novel \textit{DAG Variational Auto-Encoder} (D-VAE) to learn DAGs. Rather than using existing simultaneous message passing schemes in equation~\ref{eq:eq.mp.gnn} to encode local graph neighborhood information, they propose a novel \textit{asynchronous message passing mechanism} that allows encoding the computations (rather than structures) on DAGs. Their model allows us to build an injective mapping from the discrete space to a continuous latent space so that every DAG computation has its unique embedding in the latent space.

\section{Datasets}
\label{sec:sec6}
We categorize datasets into six diverse groups according to specific tasks: graphs for node-level tasks, graphs for graph-level tasks, graphs for generating explanations, graphs for structure learning, observed variables for inference, and observed variables for directed acyclic graph learning. \citet{dwivedi2020benchmarkgnns} and \citet{hu2020ogb} contribute great datasets for GNNs to test on the node- and graph-level tasks, provide leaderboards for fair comparisons, and evaluate the performance of different GNNs on the node- and graph-level tasks.

\subsection{Node-level Task}
We summarize selected node benchmark datasets in Table~\ref{tab:tab2}; node-level tasks include node classification and node regression.
In node-level tasks, most methods
follow full-supervised splits of train/valid/test or semi-supervised splits on benchmark datasets including citation networks, commercial networks, social networks, proteins graphs, mathematical modeling graphs: \textit{Cora, Citeseer, Pubmed,  NELL, PATTERN, CLUSTER, Reddit, ogbn-products, ogbn-proteins, ogbn-arxiv, ogbn-papers100M, ogbn-mag}. And it is required to report the number of model parameters, the total training time in second or millisecond, the number of training epochs per second or millisecond, and the average accuracy or ROC-AUC with standard deviation on the test data over at least five runtimes.

\subsection{Graph-level Task}
We summarize selected graph benchmark datasets in Table~\ref{tab:tab2}, graph-level tasks include graph classification and graph regression.
In graph-level tasks, most methods
follow full-supervised splits of train/valid/test on benchmark datasets including computer vision graphs, chemistry graphs, molecular graphs, protein graphs, social networks: \textit{MUTAG, DD, NCI1,
PROTEINS, REDDIT-BINARY, ZINC, MNIST, CIFAR10, ogbg-molhiv, ogbg-molpcba, ogbg-ppa}. And it is required to report the number of model parameters, the total training time in second or millisecond, the number of training epochs per second or millisecond, the average accuracy, AP, MAE, or ROC-AUC with standard deviation on the test data over at least five runtimes.

\subsection{GNN Explanation}
Researchers present experiments to evaluate example-level or model-level explanations for node classification and graph classification tasks on either synthetic datasets or real-world datasets, \eg{} benchmarking datasets listed in Table~\ref{tab:tab2} can also be used in GNN explanation task. Or more often,
the trust weighted signed networks, \textit{Bitcoin-Alpha} and \textit{Bitcoin-OTC}~\citep{kumar2016edge}, are two datasets for evaluating GNN explanations on node-level tasks; the computer vision graphs \textit{MNIST}~\citep{dwivedi2020benchmarkgnns}, the molecular graphs \textit{MUTAG}, and the social networks \textit{Reddit-Binary}~\citep{yanardag2015deep}, are three datasets for evaluating GNN explanations on graph-level tasks. A GNN is initially trained on these datasets to generate some representations or label distributions, then each explainable method is compared by the explanation accuracy.

\subsection{Graph Structure Learning}
Like the GNN explanation task, benchmarking datasets listed in Table~\ref{tab:tab2} can be used in graph structure learning for GNNs as well. \citet{franceschi2019learning} evaluate their model on incomplete \textit{Cora} and \textit{Citeseer} networks where they construct graphs
with missing edges by randomly sampling 25\%, 50\%, and 75\% of the edges. Authors of vGCN~\citep{elinas2020variational} also test on incomplete \textit{Cora} and \textit{Citeseer} graphs with a different setting. 

Moreover, observed variables can be used to conduct graph structure learning even if their relational graph structures are not given.
\citet{kipf2018neural} experiment their method for relational inference with three simulated systems: particles connected by springs, charged particles, and phase-coupled oscillators (Kuramoto model). \citet{franceschi2019learning} also test their model on benchmark datasets, such as \textit{Cancer, Digits}, and \textit{20news}, which are available in scikit-learn~\citep{pedregosa2011scikit}.

\subsection{Exact Inference}
In the exact inference, researchers often examine their algorithms on either synthetic graphical models or real-world graphical models. The synthetic graphical models can be Gaussian graphical models, \eg{} loopy graphs and tree graphs, or continuous non-Gaussian graphical models~\citep{fei2021generalization}. The real-world datasets include  \textit{low-density parity check (LDPC)} encoded signals where where the decoding can be done by sum/max-product belief propagation~\citep{zhang2020factor}, and the \textit{Human3.6M} dataset~\citep{ionescu2013human3} of the skeleton data for human motion prediction.

\subsection{DAG Learning}
Synthetic directed acyclic graphs are often generated for DAG learning; experiments of DAG learning only demonstrate successes on small graphical datasets. Authors in~\citep{yu2019dag} generate synthetic datasets by sampling generalized linear models, with an emphasis on nonlinear data and vector-valued data. In \citep{zhang2019d}, authors evaluate their model on two DAG optimization tasks including neural architecture search and Bayesian network structure learning.

\section{Future Directions}
\label{sec:sec7}
Though PGMs for GNNs and GNNs for PGMs have been developed in the past few years, there remain several open challenges and opportunities for exploration due to the complexity of graphs and graphical models. In this section, we discuss future directions from three perspectives.

\subsection{Benchmarks and Datasets}
Experiments for explainability, structure learning, and exact inference demonstrate successes on small graphs and graphical datasets. 
However, existing datasets for explainable GNNs, graph structure learning, inference, and DAG learning still remain limited in terms of their scale and accessibility. Graph datasets (see Table~\ref{tab:tab2}) used in classification/regression tasks are also often used in other tasks, however, each task should have its specific datasets and benchmarks for evaluation and comparison. Future experiments should consider training and testing on larger and more diverse real-world networks, as well as on broader classes of graphical models with more interesting sufficient statistics for nodes. Therefore we encourage future studies to collect more data on large diverse real-world graphs for training, and proper baselines and benchmarks for fair evaluation and comparison.

\subsection{Homphilic Graphs}
Homophily assumption on graphs is the tendency that connected nodes are more likely to share the same labels or similar features~\citep{grover2016node2vec}. When the homophily assumption fails on graph-structured data, current GNNs fail to distinguish nodes and even simple \textit{Multi-Layer Perceptrons} (MLP) can outperform GNNs by a large margin on several node classification tasks~\citep{luan2021heterophily}. The performance of GNNs can be heavily affected by varied graph structures at different homophily levels, and we will expect more accurate predictions on homophilic graphs. One possible direction is to infer or to learn homophilic graphs from observed variables. So GNNs can conduct message passing and perform training in homophilic neighbors to achieve better performance. Graph structure learning is a big topic in which PGMs benefit GNNs to learn noiseless graphs for downstream tasks. Potentially, PGM approaches could be well-adopted for generating homophilic graphs based on feature and training label similarities.

\subsection{Interpretability and Explainability}
Interpretability and explainability are two closely-related crucial components for GNN designs, and still open challenges in many real-world applications, \eg{} drug design, recommender system, and navigation satellite system.  Authors in ~\citep{yuan2020explainability} clearly distinguish interpretability and explainability in the field of GNNs. They consider a GNN to be $'$interpretable$'$ if the model itself can provide human-level understandable interpretations of its predictions, and the GNN should no longer follow some black-box mechanisms. An $'$explainable$'$ GNN implies that the model is still a black box but the predictions could potentially be understood by \textit{post hoc} explanation techniques, such as the BN approach adopted in~\citep{vu2020pgm}. How to design interpretable GNNs and how to generate explanations for GNN predictions are still under-explored. Since exact inference aims to precisely learn probability distributions and extract meaningful relationships between inputs and outputs, white-box PGMs that make interpretable and explainable message passing and predictions on graphs should be more adopted and implemented in designs of powerful GNNs.

\clearpage
\bibliography{example_paper}

\begin{thebibliography}{48}
\providecommand{\natexlab}[1]{#1}
\providecommand{\url}[1]{\texttt{#1}}
\expandafter\ifx\csname urlstyle\endcsname\relax
  \providecommand{\doi}[1]{doi: #1}\else
  \providecommand{\doi}{doi: \begingroup \urlstyle{rm}\Url}\fi

\bibitem[Bruna \& Li(2017)Bruna and Li]{bruna2017community}
Bruna, J. and Li, X.
\newblock Community detection with graph neural networks.
\newblock \emph{stat}, 1050:\penalty0 27, 2017.

\bibitem[Carlson et~al.(2010)Carlson, Betteridge, Kisiel, Settles, Hruschka,
  and Mitchell]{carlson2010toward}
Carlson, A., Betteridge, J., Kisiel, B., Settles, B., Hruschka, E.~R., and
  Mitchell, T.~M.
\newblock Toward an architecture for never-ending language learning.
\newblock In \emph{Twenty-Fourth AAAI conference on artificial intelligence},
  2010.

\bibitem[Chickering(1996)]{chickering1996learning}
Chickering, D.~M.
\newblock Learning bayesian networks is np-complete.
\newblock In \emph{Learning from data}, pp.\  121--130. Springer, 1996.

\bibitem[Dwivedi et~al.(2020)Dwivedi, Joshi, Laurent, Bengio, and
  Bresson]{dwivedi2020benchmarkgnns}
Dwivedi, V.~P., Joshi, C.~K., Laurent, T., Bengio, Y., and Bresson, X.
\newblock Benchmarking graph neural networks.
\newblock \emph{arXiv preprint arXiv:2003.00982}, 2020.

\bibitem[Elinas et~al.(2020)Elinas, Bonilla, and Tiao]{elinas2020variational}
Elinas, P., Bonilla, E.~V., and Tiao, L.
\newblock Variational inference for graph convolutional networks in the absence
  of graph data and adversarial settings.
\newblock \emph{Advances in Neural Information Processing Systems},
  33:\penalty0 18648--18660, 2020.

\bibitem[Fei \& Pitkow(2021)Fei and Pitkow]{fei2021generalization}
Fei, Y. and Pitkow, X.
\newblock Generalization of graph network inferences in higher-order
  probabilistic graphical models.
\newblock \emph{arXiv preprint arXiv:2107.05729}, 2021.

\bibitem[Franceschi et~al.(2019)Franceschi, Niepert, Pontil, and
  He]{franceschi2019learning}
Franceschi, L., Niepert, M., Pontil, M., and He, X.
\newblock Learning discrete structures for graph neural networks.
\newblock In \emph{International conference on machine learning}, pp.\
  1972--1982. PMLR, 2019.

\bibitem[Gao et~al.(2019)Gao, Pei, and Huang]{gao2019conditional}
Gao, H., Pei, J., and Huang, H.
\newblock Conditional random field enhanced graph convolutional neural
  networks.
\newblock In \emph{Proceedings of the 25th ACM SIGKDD International Conference
  on Knowledge Discovery \& Data Mining}, pp.\  276--284, 2019.

\bibitem[Gatterbauer(2017)]{gatterbauer2017linearization}
Gatterbauer, W.
\newblock The linearization of belief propagation on pairwise markov random
  fields.
\newblock In \emph{Proceedings of the AAAI Conference on Artificial
  Intelligence}, volume~31, 2017.

\bibitem[Gilmer et~al.(2017)Gilmer, Schoenholz, Riley, Vinyals, and
  Dahl]{gilmer2017neural}
Gilmer, J., Schoenholz, S.~S., Riley, P.~F., Vinyals, O., and Dahl, G.~E.
\newblock Neural message passing for quantum chemistry.
\newblock In \emph{Proceedings of the 34th International Conference on Machine
  Learning-Volume 70}, pp.\  1263--1272. JMLR. org, 2017.

\bibitem[Grover \& Leskovec(2016)Grover and Leskovec]{grover2016node2vec}
Grover, A. and Leskovec, J.
\newblock node2vec: Scalable feature learning for networks.
\newblock In \emph{Proceedings of the 22nd ACM SIGKDD international conference
  on Knowledge discovery and data mining}, pp.\  855--864. ACM, 2016.

\bibitem[Hamilton(2020)]{hamilton2020graph}
Hamilton, W.~L.
\newblock Graph representation learning.
\newblock \emph{Synthesis Lectures on Artifical Intelligence and Machine
  Learning}, 14\penalty0 (3):\penalty0 1--159, 2020.

\bibitem[Hamilton et~al.(2017)Hamilton, Ying, and
  Leskovec]{hamilton2017inductive}
Hamilton, W.~L., Ying, R., and Leskovec, J.
\newblock Inductive representation learning on large graphs.
\newblock \emph{arXiv}, abs/1706.02216, 2017.
\newblock URL \url{http://arxiv.org/abs/1706.02216}.

\bibitem[Hu et~al.(2020)Hu, Fey, Zitnik, Dong, Ren, Liu, Catasta, and
  Leskovec]{hu2020ogb}
Hu, W., Fey, M., Zitnik, M., Dong, Y., Ren, H., Liu, B., Catasta, M., and
  Leskovec, J.
\newblock Open graph benchmark: Datasets for machine learning on graphs.
\newblock \emph{arXiv preprint arXiv:2005.00687}, 2020.

\bibitem[Hua et~al.(2022)Hua, Rabusseau, and Tang]{hua2022high}
Hua, C., Rabusseau, G., and Tang, J.
\newblock High-order pooling for graph neural networks with tensor
  decomposition.
\newblock \emph{arXiv preprint arXiv:2205.11691}, 2022.

\bibitem[Ionescu et~al.(2013)Ionescu, Papava, Olaru, and
  Sminchisescu]{ionescu2013human3}
Ionescu, C., Papava, D., Olaru, V., and Sminchisescu, C.
\newblock Human3. 6m: Large scale datasets and predictive methods for 3d human
  sensing in natural environments.
\newblock \emph{IEEE transactions on pattern analysis and machine
  intelligence}, 36\penalty0 (7):\penalty0 1325--1339, 2013.

\bibitem[Kipf et~al.(2018)Kipf, Fetaya, Wang, Welling, and
  Zemel]{kipf2018neural}
Kipf, T., Fetaya, E., Wang, K.-C., Welling, M., and Zemel, R.
\newblock Neural relational inference for interacting systems.
\newblock In \emph{International Conference on Machine Learning}, pp.\
  2688--2697. PMLR, 2018.

\bibitem[Kipf \& Welling(2016)Kipf and Welling]{kipf2016classification}
Kipf, T.~N. and Welling, M.
\newblock Semi-supervised classification with graph convolutional networks.
\newblock \emph{arXiv}, abs/1609.02907, 2016.
\newblock URL \url{http://arxiv.org/abs/1609.02907}.

\bibitem[Koller \& Friedman(2009)Koller and Friedman]{koller2009probabilistic}
Koller, D. and Friedman, N.
\newblock \emph{Probabilistic graphical models: principles and techniques}.
\newblock MIT press, 2009.

\bibitem[Kumar et~al.(2016)Kumar, Spezzano, Subrahmanian, and
  Faloutsos]{kumar2016edge}
Kumar, S., Spezzano, F., Subrahmanian, V., and Faloutsos, C.
\newblock Edge weight prediction in weighted signed networks.
\newblock In \emph{2016 IEEE 16th International Conference on Data Mining
  (ICDM)}, pp.\  221--230. IEEE, 2016.

\bibitem[Li et~al.(2018)Li, Vinyals, Dyer, Pascanu, and
  Battaglia]{li2018learning}
Li, Y., Vinyals, O., Dyer, C., Pascanu, R., and Battaglia, P.
\newblock Learning deep generative models of graphs.
\newblock \emph{arXiv preprint arXiv:1803.03324}, 2018.

\bibitem[Liu et~al.(2019)Liu, Gao, Khan, Qi, and Guan]{liu2019graph}
Liu, K., Gao, L., Khan, N.~M., Qi, L., and Guan, L.
\newblock Graph convolutional networks-hidden conditional random field model
  for skeleton-based action recognition.
\newblock In \emph{2019 IEEE International Symposium on Multimedia (ISM)}, pp.\
   25--256. IEEE, 2019.

\bibitem[Luan et~al.(2020)Luan, Zhao, Hua, Chang, and Precup]{luan2020complete}
Luan, S., Zhao, M., Hua, C., Chang, X.-W., and Precup, D.
\newblock Complete the missing half: Augmenting aggregation filtering with
  diversification for graph convolutional networks.
\newblock \emph{arXiv preprint arXiv:2008.08844}, 2020.

\bibitem[Luan et~al.(2021)Luan, Hua, Lu, Zhu, Zhao, Zhang, Chang, and
  Precup]{luan2021heterophily}
Luan, S., Hua, C., Lu, Q., Zhu, J., Zhao, M., Zhang, S., Chang, X.-W., and
  Precup, D.
\newblock Is heterophily a real nightmare for graph neural networks to do node
  classification?
\newblock \emph{arXiv preprint arXiv:2109.05641}, 2021.

\bibitem[Luan et~al.(2022)Luan, Hua, Lu, Zhu, Zhao, Zhang, Chang, and
  Precup]{luan2022revisiting}
Luan, S., Hua, C., Lu, Q., Zhu, J., Zhao, M., Zhang, S., Chang, X.-W., and
  Precup, D.
\newblock Revisiting heterophily for graph neural networks.
\newblock \emph{arXiv preprint arXiv:2210.07606}, 2022.

\bibitem[Ma et~al.(2018)Ma, Xiao, Shang, and Sun]{ma2018cgnf}
Ma, T., Xiao, C., Shang, J., and Sun, J.
\newblock Cgnf: Conditional graph neural fields.
\newblock 2018.

\bibitem[Murphy et~al.(2013)Murphy, Weiss, and Jordan]{murphy2013loopy}
Murphy, K., Weiss, Y., and Jordan, M.~I.
\newblock Loopy belief propagation for approximate inference: An empirical
  study.
\newblock \emph{arXiv preprint arXiv:1301.6725}, 2013.

\bibitem[Pedregosa et~al.(2011)Pedregosa, Varoquaux, Gramfort, Michel, Thirion,
  Grisel, Blondel, Prettenhofer, Weiss, Dubourg, et~al.]{pedregosa2011scikit}
Pedregosa, F., Varoquaux, G., Gramfort, A., Michel, V., Thirion, B., Grisel,
  O., Blondel, M., Prettenhofer, P., Weiss, R., Dubourg, V., et~al.
\newblock Scikit-learn: Machine learning in python.
\newblock \emph{the Journal of machine Learning research}, 12:\penalty0
  2825--2830, 2011.

\bibitem[Qu et~al.(2019)Qu, Bengio, and Tang]{qu2019gmnn}
Qu, M., Bengio, Y., and Tang, J.
\newblock Gmnn: Graph markov neural networks.
\newblock In \emph{International conference on machine learning}, pp.\
  5241--5250. PMLR, 2019.

\bibitem[Qu et~al.(2021)Qu, Cai, and Tang]{qu2021neural}
Qu, M., Cai, H., and Tang, J.
\newblock Neural structured prediction for inductive node classification.
\newblock In \emph{International Conference on Learning Representations}, 2021.

\bibitem[Satorras \& Welling(2021)Satorras and Welling]{satorras2021neural}
Satorras, V.~G. and Welling, M.
\newblock Neural enhanced belief propagation on factor graphs.
\newblock In \emph{International Conference on Artificial Intelligence and
  Statistics}, pp.\  685--693. PMLR, 2021.

\bibitem[Sen et~al.(2008)Sen, Namata, Bilgic, Getoor, Galligher, and
  Eliassi-Rad]{sen2008collective}
Sen, P., Namata, G., Bilgic, M., Getoor, L., Galligher, B., and Eliassi-Rad, T.
\newblock Collective classification in network data.
\newblock \emph{AI magazine}, 29\penalty0 (3):\penalty0 93--93, 2008.

\bibitem[Tang et~al.(2021)Tang, Chen, Bai, Liu, Ge, and Ouyang]{tang2021mutual}
Tang, S., Chen, D., Bai, L., Liu, K., Ge, Y., and Ouyang, W.
\newblock Mutual crf-gnn for few-shot learning.
\newblock In \emph{Proceedings of the IEEE/CVF Conference on Computer Vision
  and Pattern Recognition}, pp.\  2329--2339, 2021.

\bibitem[Thost \& Chen(2021)Thost and Chen]{thost2021directed}
Thost, V. and Chen, J.
\newblock Directed acyclic graph neural networks.
\newblock \emph{arXiv preprint arXiv:2101.07965}, 2021.

\bibitem[Velickovic et~al.(2017)Velickovic, Cucurull, Casanova, Romero, Lio,
  and Bengio]{velivckovic2017attention}
Velickovic, P., Cucurull, G., Casanova, A., Romero, A., Lio, P., and Bengio, Y.
\newblock Graph attention networks.
\newblock \emph{arXiv}, abs/1710.10903, 2017.

\bibitem[Vu \& Thai(2020)Vu and Thai]{vu2020pgm}
Vu, M. and Thai, M.~T.
\newblock Pgm-explainer: Probabilistic graphical model explanations for graph
  neural networks.
\newblock \emph{Advances in neural information processing systems},
  33:\penalty0 12225--12235, 2020.

\bibitem[Wang et~al.(2021)Wang, Jia, and Gong]{wang2021semi}
Wang, B., Jia, J., and Gong, N.~Z.
\newblock Semi-supervised node classification on graphs: Markov random fields
  vs. graph neural networks.
\newblock In \emph{Proceedings of the AAAI Conference on Artificial
  Intelligence}, volume~35, pp.\  10093--10101, 2021.

\bibitem[Wang et~al.(2020)Wang, Li, Long, Zhang, Song, and
  Shi]{wang2020learning}
Wang, J., Li, Z., Long, Q., Zhang, W., Song, G., and Shi, C.
\newblock Learning node representations from noisy graph structures.
\newblock In \emph{2020 IEEE International Conference on Data Mining (ICDM)},
  pp.\  1310--1315. IEEE, 2020.

\bibitem[Yanardag \& Vishwanathan(2015)Yanardag and
  Vishwanathan]{yanardag2015deep}
Yanardag, P. and Vishwanathan, S.
\newblock Deep graph kernels.
\newblock In \emph{Proceedings of the 21th ACM SIGKDD international conference
  on knowledge discovery and data mining}, pp.\  1365--1374, 2015.

\bibitem[Ying et~al.(2019)Ying, Bourgeois, You, Zitnik, and
  Leskovec]{ying2019gnnexplainer}
Ying, Z., Bourgeois, D., You, J., Zitnik, M., and Leskovec, J.
\newblock Gnnexplainer: Generating explanations for graph neural networks.
\newblock \emph{Advances in neural information processing systems}, 32, 2019.

\bibitem[Yoon et~al.(2019)Yoon, Liao, Xiong, Zhang, Fetaya, Urtasun, Zemel, and
  Pitkow]{yoon2019inference}
Yoon, K., Liao, R., Xiong, Y., Zhang, L., Fetaya, E., Urtasun, R., Zemel, R.,
  and Pitkow, X.
\newblock Inference in probabilistic graphical models by graph neural networks.
\newblock In \emph{2019 53rd Asilomar Conference on Signals, Systems, and
  Computers}, pp.\  868--875. IEEE, 2019.

\bibitem[Yu et~al.(2019)Yu, Chen, Gao, and Yu]{yu2019dag}
Yu, Y., Chen, J., Gao, T., and Yu, M.
\newblock Dag-gnn: Dag structure learning with graph neural networks.
\newblock In \emph{International Conference on Machine Learning}, pp.\
  7154--7163. PMLR, 2019.

\bibitem[Yuan \& Ji(2020)Yuan and Ji]{yuan2020structpool}
Yuan, H. and Ji, S.
\newblock Structpool: Structured graph pooling via conditional random fields.
\newblock In \emph{Proceedings of the 8th International Conference on Learning
  Representations}, 2020.

\bibitem[Yuan et~al.(2020)Yuan, Yu, Gui, and Ji]{yuan2020explainability}
Yuan, H., Yu, H., Gui, S., and Ji, S.
\newblock Explainability in graph neural networks: A taxonomic survey.
\newblock \emph{arXiv preprint arXiv:2012.15445}, 2020.

\bibitem[Zhang et~al.(2019{\natexlab{a}})Zhang, Jiang, Cui, Garnett, and
  Chen]{zhang2019d}
Zhang, M., Jiang, S., Cui, Z., Garnett, R., and Chen, Y.
\newblock D-vae: A variational autoencoder for directed acyclic graphs.
\newblock \emph{Advances in Neural Information Processing Systems}, 32,
  2019{\natexlab{a}}.

\bibitem[Zhang et~al.(2019{\natexlab{b}})Zhang, Pal, Coates, and
  Ustebay]{zhang2019bayesian}
Zhang, Y., Pal, S., Coates, M., and Ustebay, D.
\newblock Bayesian graph convolutional neural networks for semi-supervised
  classification.
\newblock In \emph{Proceedings of the AAAI conference on artificial
  intelligence}, volume~33, pp.\  5829--5836, 2019{\natexlab{b}}.

\bibitem[Zhang et~al.(2020)Zhang, Wu, and Lee]{zhang2020factor}
Zhang, Z., Wu, F., and Lee, W.~S.
\newblock Factor graph neural networks.
\newblock \emph{Advances in Neural Information Processing Systems},
  33:\penalty0 8577--8587, 2020.

\bibitem[Zhu et~al.(2021)Zhu, Xu, Zhang, Liu, Wu, and Wang]{zhu2021deep}
Zhu, Y., Xu, W., Zhang, J., Liu, Q., Wu, S., and Wang, L.
\newblock Deep graph structure learning for robust representations: A survey.
\newblock \emph{arXiv preprint arXiv:2103.03036}, 2021.

\end{thebibliography}
\bibliographystyle{icml2022}

%%%%%%%%%%%%%%%%%%%%%%%%%%%%%%%%%%%%%%%%%%%%%%%%%%%%%%%%%%%%%%%%%%%%%%%%%%%%%%%
%%%%%%%%%%%%%%%%%%%%%%%%%%%%%%%%%%%%%%%%%%%%%%%%%%%%%%%%%%%%%%%%%%%%%%%%%%%%%%%
% APPENDIX
%%%%%%%%%%%%%%%%%%%%%%%%%%%%%%%%%%%%%%%%%%%%%%%%%%%%%%%%%%%%%%%%%%%%%%%%%%%%%%%
%%%%%%%%%%%%%%%%%%%%%%%%%%%%%%%%%%%%%%%%%%%%%%%%%%%%%%%%%%%%%%%%%%%%%%%%%%%%%%%
\newpage
\appendix
\onecolumn
\appendix

\end{document}